\documentclass[pmlr,twocolumn,10pt]{jmlr}
\usepackage{longtable}
\usepackage{booktabs}
\usepackage{multirow}
\usepackage{multicol}
\usepackage{colortbl}
\usepackage{caption}
\usepackage{adjustbox}
\usepackage{xurl}
\usepackage{cleveref}
\usepackage{siunitx}

\usepackage[switch]{lineno}

\theorembodyfont{\upshape}
\theoremheaderfont{\scshape}
\theorempostheader{:}
\theoremsep{\newline}

\jmlrvolume{287}
\jmlryear{2025}
\jmlrsubmitted{} %LEAVE UNSET
\jmlrpublished{} %LEAVE UNSET
\jmlrworkshop{Conference on Health, Inference, and Learning (CHIL) 2025} % W&CP title

\title{Uncertainty Quantification for Machine Learning in Healthcare:\\ A Survey}

\author{%
 \Name{L. Julián Lechuga López $^{1,2}$} \Email{leopoldo.lechuga@nyu.edu}\\
 \Name{Shaza Elsharief $^2$} \Email{se1525@nyu.edu}\\
 \Name{Dhiyaa Al Jorf $^2$} \Email{da2863@nyu.edu}\\
 \Name{Firas Darwish $^2$} \Email{fbd2014@nyu.edu}\\
 \Name{Congbo Ma $^2$} \Email{cm7196@nyu.edu}\\
 \Name{Farah E. Shamout $^{1,2}$} \Email{farah.shamout@nyu.edu}\\
 \addr New York University $^1$, New York University Abu Dhabi $^2$
}

\begin{document}

\maketitle

\begin{abstract}
%  Currently sitting at 400 words, so we can extract the most useful info.
% SCARIC
% S: The SITUATION, background and related work in your research area
Uncertainty Quantification (UQ) is pivotal in enhancing the robustness, reliability, and interpretability of Machine Learning (ML) systems for healthcare, optimizing resources and improving patient care.
% Particularly in healthcare, ML-driven systems promise to completely transform the clinical landscape of automated patient care, optimizing resources and further improving diagnostics and prognosis. 
% C: The main CHALLENGE(s) in your field that need to be urgently addressed
Despite the emergence of ML-based clinical decision support tools, the lack of principled quantification of uncertainty in ML models remains a major challenge.
% Since uncertainty is a ubiquitous phenomenon, it can generate and propagate across the whole ML pipeline.
% This requires tailored solutions that address uncertainty at each stage of the ML pipeline, adapting it to the constraints of the medical domain.
% Current review studies have a narrow focus on analyzing the state-of-the-art uncertainty quantification applied in healthcare for specific domains, without providing a systematic evaluation of method efficacy across different stages of model development.
%
Current reviews have a narrow focus on analyzing the state-of-the-art UQ in specific healthcare domains without systematically evaluating method efficacy across different stages of model development, and despite a growing body of research, its implementation in healthcare applications remains limited.
%
% A: The methodology and ACTIONS needed to be taken to address the challenge
Therefore, in this survey, we provide a comprehensive analysis of current UQ in healthcare, offering an informed framework that highlights how different methods can be integrated into each stage of the ML pipeline including data processing, training and evaluation.
% We extracted information from peer-reviewed studies in the areas of uncertainty quantification (non-healthcare) and uncertainty quantification in healthcare to analyze the most popular methods used in healthcare and novel approaches from other domains that hold potential for future adoption in the medical context. 
% R: The key RESULTS, findings or measurable outcomes of your research
% Our findings show that despite a growing body of studies focused on uncertainty quantification, their implementation in healthcare applications is limited.
We also highlight the most popular methods used in healthcare and novel approaches from other domains that hold potential for future adoption in the medical context.
% applications mainly apply well known UQ approaches rather than developing state-of-the-art solutions specifically tailored for healthcare.
% In addition, most studies in healthcare are limited to clinical imaging datasets, which hinders the development of robust medical applications that can adapt to the multimodal requirements of patient care.
% I: The academic and socioeconomic IMPACT of your research
%
We expect this study will provide a clear overview of the challenges and opportunities of implementing UQ in the ML pipeline for healthcare, guiding researchers and practitioners in selecting suitable techniques to enhance the reliability, safety and trust from patients and clinicians on ML-driven healthcare solutions. 

% C: A strong CALL-TO-ACTION that motivates your readers to explore further directions
% ??

\end{abstract}
%%%%%%%%%%%%%%%%%%%%%%%%%%%%%%%%%%%%%%%%%%%%%%%%%%%%%%%%%%%%%%%%%%%%%%%%%%%

\paragraph*{Data and Code Availability}
This literature review does not rely on any specific dataset, as it synthesizes findings from existing research on UQ in healthcare. 
No new data was generated, and no code was developed or available for sharing.

\paragraph*{Institutional Review Board (IRB)}
This literature review on UQ in healthcare does not involve human subjects, so IRB approval was not required.

\section{Introduction}
\label{sec:introduction}

\begin{figure*}[t!]
    \centering
    \includegraphics[width=0.8\textwidth]{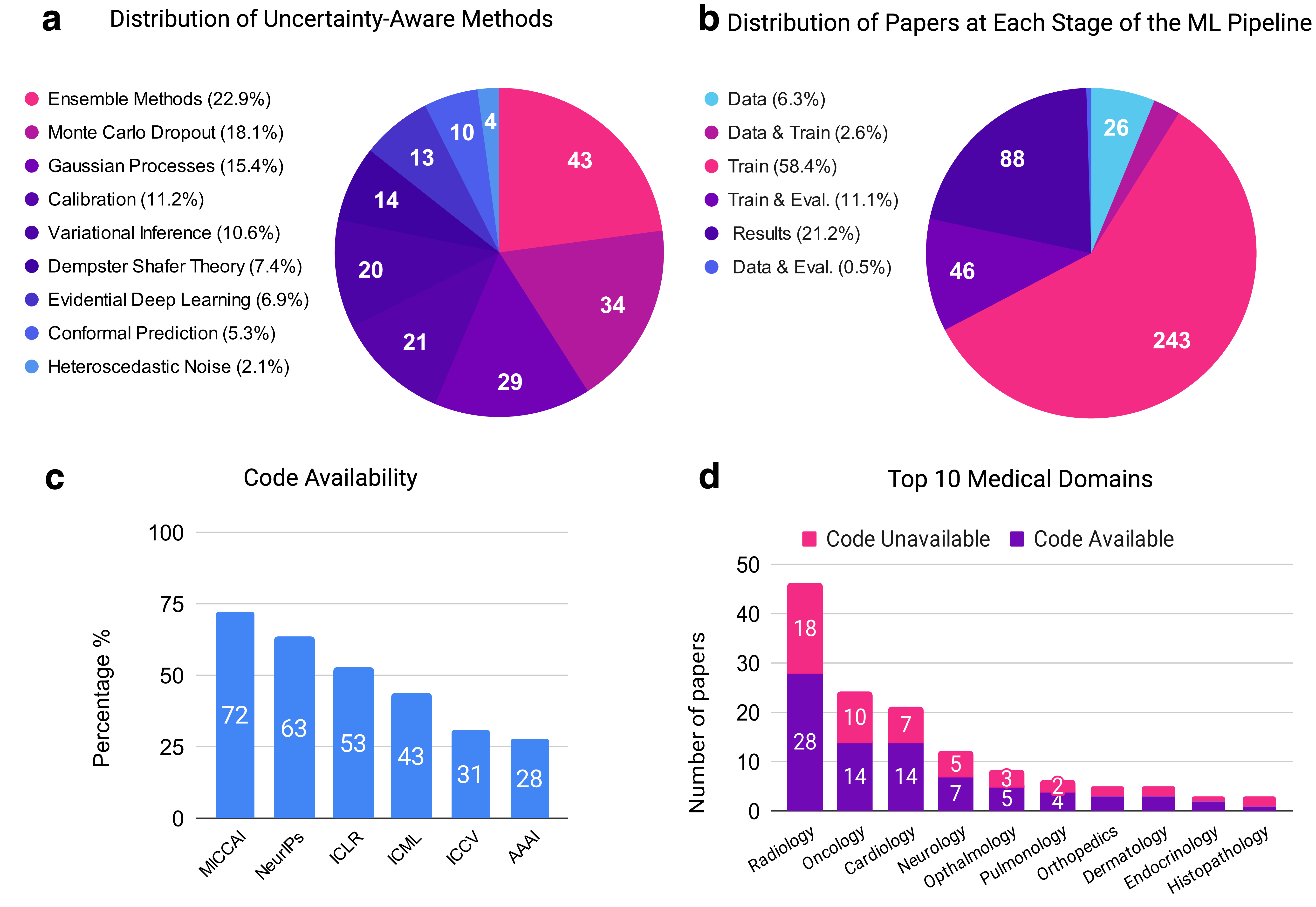}
    \caption{\textbf{Overview of distribution and characteristics of reviewed papers.}
    \textbf{(a)} Prevalence of different uncertainty quantification methods across the surveyed papers.
    \textbf{(b)} Distribution of studies according to the machine learning pipeline stages: data processing, model training, and evaluation.
    \textbf{(c)} Code availability rates across papers published in various conferences and journals.
    \textbf{(d)} 
    Medical domains represented in the reviewed studies, alongside their corresponding code availability.
    }
    \label{fig:figure_1}
\end{figure*}

% P1: Introduction of ML in healthcare
Machine Learning (ML) is revolutionizing healthcare by enhancing diagnostic performance, personalizing treatment plans, optimizing hospital operations, and accelerating drug discovery, ultimately leading to improved patient outcomes and more efficient and safe medical practices \citep{Revolutionizing_Healthcare}.
Many systems have been developed to support clinical diagnosis \citep{browning_uncertainty_2021}, from analyzing medical images for anomaly detection \citep{Zhou_Review}, to providing personalized treatment plans based on patient-specific physiological and genetic characteristics \citep{improving_image_based_precision_medicine}.

% P2: Relevance of UQ in healthcare
However, due to the safety-critical nature of clinical practice, the development of trustworthy and deployable ML in healthcare requires the implementation of robust Uncertainty Quantification (UQ) \citep{begoli_need_2019, gruber_sources_2023}. 
Variations in real-world clinical environments affect the performance of predictive systems and introduce uncertainty at different stages of the ML pipeline: data noise and distribution drift, bias and miscalibration of model parameters, or evaluation of the model in an out-of-distribution scenario, such as deployment in a different hospital \citep{OOD_medical}.
By complementing AI-driven healthcare systems with assessments of the uncertainty in their predictions, methods can help better explain whether errors can be attributed to randomness and noise or whether they are due to design choices made during model training and development. 

% P3: Benefits of uncertainty quantification ML healthcare for clinicians and patients
Furthermore, the development of UQ in ML for healthcare can improve confidence in the adoption of these tools by clinicians, patients, and institutions alike \citep{kurz2022uncertainty}.
Robust and informative UQ diagnostics can help clinicians focus their attention on specific details of patient data, and distinguish between predictions made with high confidence and those with substantial ambiguity, which allows for better risk management \citep{ren2023uncertaintyinformedmutuallearningjoint}. 
For patients, understanding the degree of confidence associated with their personalized predictions allows them to weigh the risks and benefits of different interventions, reduces the chances of inappropriate or invasive treatment, and improves their trust in the model \citep{improving_image_based_precision_medicine}. 
This transparency is vital for healthcare institutions, as it leads to better informed decision-making, optimization of operations, reduction of misdiagnoses and incorrect treatment recommendations, and most importantly, improved patient care \citep{Zhou_Review}.

%P4: Uncertainty gaps related to AI in healthcare and relationship to the ML pipeline
Unfortunately, real-world implementation of UQ models in healthcare is still hampered by the limited development of tailored solutions for improving these models \citep{ovadia2019can, kompa_second_2021}.
Limited underlying theory on how to best adapt predictive uncertainty methods in clinical tasks shows that the use of UQ in clinical applications is not common practice \citep{begoli_need_2019, lambert_trustworthy_2024}.
In addition, since uncertainty in healthcare applications originates at different stages, it needs to be analyzed from the perspective of  the ML pipeline: data processing, model training, and model evaluation.
Therefore, there is a crucial need to design and develop Uncertainty Quantification for Machine Learning in Healthcare (UQML4H) to enable the implementation of trustworthy systems that are adapted to robustly mitigate uncertainty during the full ML design lifecycle.
Our main goal is to provide a comprehensive overview of State-of-the-Art (SOTA) UQ methods in healthcare, clinical datasets, tasks and domains, and to encourage the development of new UQ methodologies that tackle specific challenges relevant to the nature of each medical domain.

\subsection{Motivation}
Most existing reviews in UQ cover a wide range of applications and domains, with narrow focus on healthcare.
These studies highlighting UQ applications in healthcare primarily focus on a single type of data modality (i.e., medical imaging) \citep{huang_review_2024} or a specific clinical task \citep{BARBANO2022601}.
However, no previous studies pay particular attention to analyzing UQ methods from an ML pipeline point of view. 
Our work is intended to bridge SOTA research in UQ and tailored clinical applications, with an emphasis on analyzing each phase of model development.
Compared to existing work, our survey has four main contributions: 

\begin{enumerate}
\setlength\itemsep{0.5mm}
    \item We focus on recent SOTA literature on UQ from both medical and nonmedical domains published in the last four years, capturing extensive applications in healthcare (e.g., diagnosis, decision-support systems, etc.) and methodological advances (e.g., theory, algorithms, optimization).
    
    \item We distinguish and analyze UQ methods at each stage of the ML pipeline: data processing (e.g., collection, labeling, alignment), model training (e.g., architecture, tuning, loss design), and evaluation (e.g., inference, metrics, calibration).
    
    \item We present a comprehensive taxonomy of UQ methods categorized by domain, dataset, and task, providing a practical reference for domain-specific applications in healthcare.
    
    \item We connect current methodological advances with practical deployment considerations of UQ in healthcare, identifying key gaps and outlining future research directions and open challenges.

\end{enumerate}

\subsection*{Scope of the Review}

To this end, the reviewed papers (overview provided in Figure \ref{fig:figure_1}) were selected using the following criteria:

\begin{itemize}
\setlength\itemsep{0.5mm}
    \item Recently published peer-reviewed work, (i.e., publication year $\geq 2020$), and earlier seminal papers in UQ.
    \item Articles from top-tier AI conferences and medical journals that focus on the application or development of UQ at any stage of the ML pipeline.
    \item Key information extracted: UQ methods, code availability, clinical datasets and tasks, and specific healthcare domains targeted (e.g., oncology, radiology, cardiology, etc.). 
\end{itemize} 

In \Cref{sec:ml_pipeline}, we present our main findings regarding the applications of UQ in healthcare across each stage of the ML pipeline, discussing both current applications and emerging opportunities. 
The section concludes with a comparative analysis of different UQ techniques and applications in healthcare, highlighting the strengths and weaknesses of seven representative methods.
\Cref{sec:future}, presents a thorough discussion of open challenges and future research directions, touching on details of deployment, fairness, regulation, evaluation benchmarks, and a roadmap towards safe UQ implementation in healthcare.
Concluding remarks are provided in \Cref{sec:conclusion}. 

In \Cref{sec:foundations_uq}, we provide a concise overview of key UQ methodologies and identify application domains closely linked to current SOTA developments in machine learning. 
Furthermore, \Cref{app:medical_datasets} (Table \ref{table:table_datasets}) presents {detailed information on 94 open-source clinical datasets}, grouped by medical domain, currently used in the development and evaluation of UQ methods. 
\Cref{app:appendix_healthcare_studies} (\Cref{table:ML_pipeline_healthcare1})  summarizes the main characteristics of the studies reviewed in this survey, organized by ML pipeline stage and clinical context. 

We anticipate that this comprehensive collection of resources will serve as a valuable reference for researchers and practitioners aiming to advance UQ applications in healthcare.

\begin{figure*}[t!]
    \centering
    \includegraphics[width=0.92\textwidth]{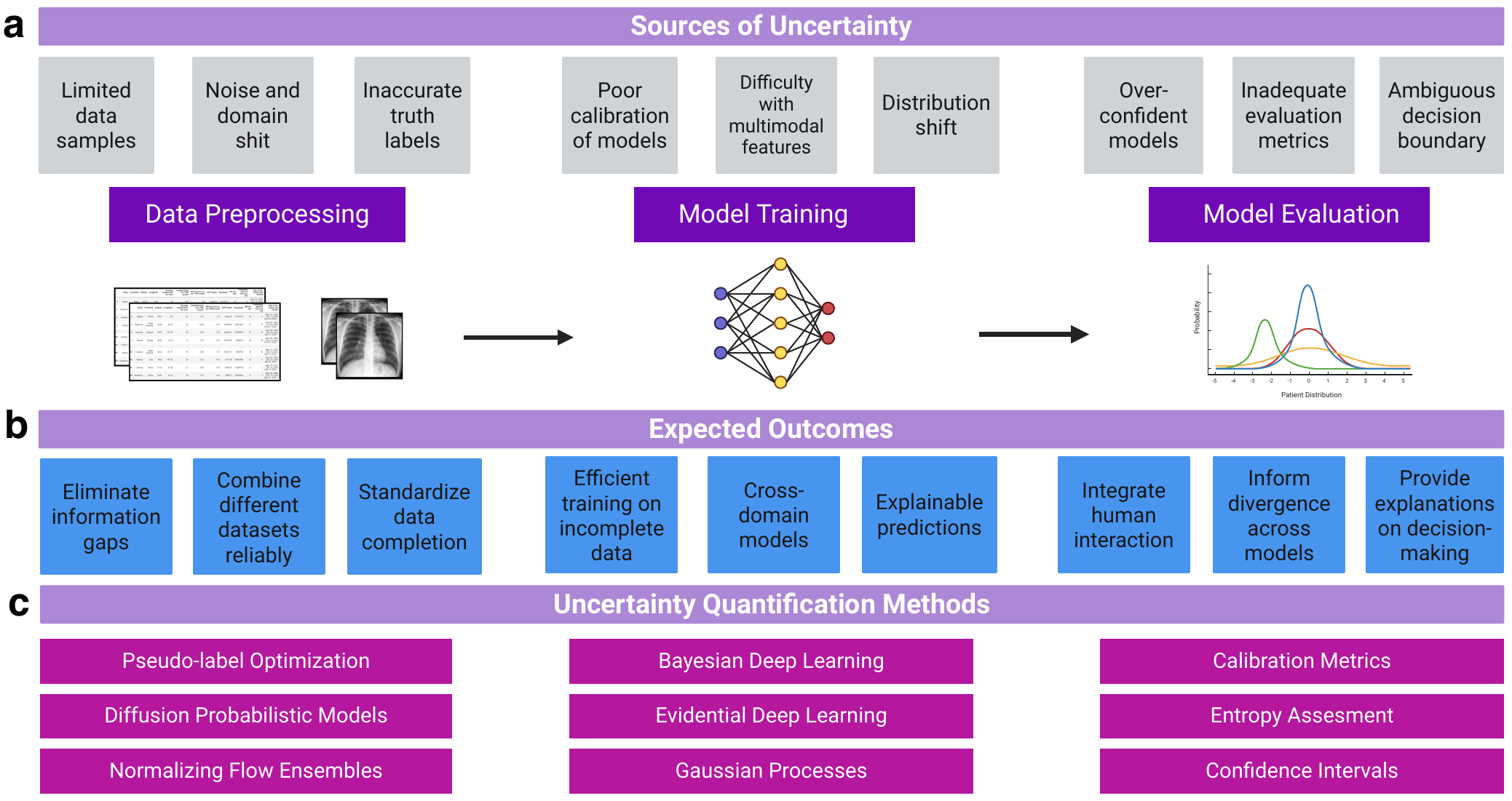}
    \caption{\textbf{UQ in the Clinical Machine Learning Pipeline.} 
    \textbf{(a)} Key sources of uncertainty identified at each stage of the pipeline.
    \textbf{(b)} Expected outcomes of implementing UQ methods for clinical tasks.
    \textbf{(c)} Relevant UQ techniques applied during data processing, model training, and evaluation.
    }
    \label{fig:figure_3}
\end{figure*}

\section{Uncertainty Quantification in Healthcare: Applications Across the Machine Learning Pipeline}
\label{sec:ml_pipeline}

This section synthesizes key findings on the use of UQ methods in healthcare across the ML pipeline, spanning data preprocessing, model training, and evaluation. 
We highlight methods applied across diverse clinical tasks, medical domains, and datasets, and point to emerging trends from other fields that may shape future developments. 
Figure \ref{fig:figure_3} summarizes the main insights, illustrating sources of uncertainty, expected outcomes, and representative UQ techniques at each pipeline stage. 
Table \ref{table:ML_pipeline_healthcare1} (Appendix \ref{app:appendix_healthcare_studies}) provides detailed information on all reviewed studies discussed in this section. 
This synthesis offers a structured overview of how UQ is currently integrated into healthcare ML workflows.

%  DATA PREPROCESSING
\subsection{Data Preprocessing} 
% \vspace{-1.5mm}
The data preprocessing stage is fundamental for ML modeling, addressing challenges like noise, imbalance, and incompleteness that increase uncertainty. 
UQ methods at this stage focus on enhancing the quality and reliability of inputs for subsequent stages.
Techniques include correcting label noise in imbalanced datasets, UQ pseudo-labeling, and modeling uncertainty with normalizing flows.

% PARAGRAPH ON HEALTHCARE APPLICATIONS
\subsubsection{Applications} 

Although methods for data preprocessing in healthcare are limited, and the boundary between preprocessing and early training stages may be ambiguous, several studies have proposed UQ methods to enhance data reliability before training.
For instance, \cite{angelopoulos2022image} made use of a distribution-free UQ method for image-to-image regression for MRI and microscopy imaging, providing pixel-wise uncertainty intervals to preprocess the input data with formal statistical guarantees addressing uncertainty.
Similarly, \cite{li2022ultra} improved the quality of data annotations using label probability distributions for tumor cellularity assessment in breast cancer histological images.
\cite{das_confidence-guided_2024} proposed AnoMed, a semi-supervised confidence guided pseudo-label optimizer, to capture anatomical structures and underlying representations in chest X-rays.
In multichannel brain MRI, \cite{tanno_uncertainty_2019} introduced a method to decompose predictive uncertainty and quantify the effects of intrinsic and parameter uncertainty of data. 
Working on echocardiography images, \cite{gu_reliable_2024} presented Re-Training for Uncertainty (RT4U), a data-centric method to introduce uncertainty to weakly informative inputs in the data. Using diffusion probabilistic modeling, \cite{oh_uncertainty-aware_2024} used data augmentation and synthesis to address domain shift issues, while \citet{khader2023denoising, adib2023synthetic, iuliano2024denoising} focused on generating high-quality synthetic data and assessed the association between original and synthetic data for 3D medical images, MRI and malaria images, and ECG data.

Beyond specific applications, the effectiveness of UQ methods at the data processing stage is also shaped by the type of data modality and its resolution, which introduce distinct sources of uncertainty and influence methodological suitability.
For instance, uncertainty in MRI may arise from acquisition parameters or reconstruction artifacts \citep{edupuganti_uncertainty_2021, zhao_lost_2024}, X-ray imaging can suffer from low contrast and dose variability \citep{cxad,gong_patient-specific_2023}, and EHR data is prone to missing values and documentation inconsistencies across institutions \citep{ehr_1, ehr_2}.
In time-series data, higher sampling rates can capture finer dynamics but risk overfitting to noise; lower rates reduce noise sensitivity but may miss short-term patterns \citep{puri2022forecasting, folgado_explainability_2023}.

To address this, UQ methods should be tailored to modality-specific characteristics: Bayesian Neural Networks (BNNs) are well-suited for pixel-level uncertainty in imaging, while deep ensembles are more effective for structured EHR data \citep{caldeira_deeply_2020, peng_bayesian_2020}. 
Multi-scale UQ approaches and ensemble techniques can also help strike a balance between sensitivity and robustness \citep{vranken_uncertainty_2021, hamedani-karazmoudehfar_breast_2023}.
Data resolution also plays a pivotal role, as high resolution imaging scans offer greater detail but introduce noise and computational overhead, whereas low-resolution data may omit clinically relevant features, affecting the reliability of UQ estimates \citep{nehme2023uncertainty, fu2025cam}.
Finally, in multimodal settings, such as those involving imaging and EHR, heterogeneous uncertainty sources must be jointly modeled with techniques like hierarchical Bayesian models. Uncertainty-weighted fusion can also account for these modality-specific variances, enhancing the overall reliability of clinical predictions \citep{fu2025cam}.

\subsubsection{Opportunities}
% While existing studies have demonstrated the role of UQ in improving data reliability during preprocessing, ongoing research continues to explore novel strategies for handling uncertainty at this stage. 

\paragraph{Label Handling.} This focuses on addressing noisy and imbalanced datasets through adaptive labeling techniques. 
Relevant examples include uncertainty quantification pseudo-labeling strategies for semi-supervised learning that filter unreliable pseudo-labels to enhance model robustness \citep{rizve_defense_2021, yan_unsupervised_2022} and Uncertainty Correction of Labels (UCL) proposed by \cite{huang_uncertainty-aware_2022} to identify and correct mislabeled data. 

\paragraph{Noise Reduction and Augmentation.} These methods attenuate data variance and calibrate noise levels, preventing spurious uncertainty signals. 
Techniques employing aleatoric uncertainty estimation and UQ augmentation can help refine domain adaptation by aligning source and target distributions \citep{yan_unsupervised_2022, zhang_one_2024}.

\paragraph{Input Transformations.} Input transformation methods are often used to improve uncertainty estimates. 
For example, normalizing flow ensembles and the use of confidence score calibration during preprocessing enables accurate representation of both aleatoric and epistemic uncertainties \citep{berry_normalizing_2023, yang2024efficient}.
These studies demonstrate the meaningful integration of UQ into the data preprocessing stage to mitigate the impact of poor quality data, enhance reliability, improve annotation quality, and mitigate uncertainties before training begins.
%%%%%%%%%%%%%%%%%%%%%%%%%%%%%%%%%%%%%%%%%%%%%%%%

% MODEL TRAINING
\subsection{Model Training}

This stage is crucial for integrating UQ, as it impacts generalization and reliability of predictions. 
UQ methods at this stage focus on capturing both epistemic and aleatoric uncertainty to enhance model robustness, reduce overfitting, and improve predictive performance under uncertainty.

% PARAGRAPH ON HEALTHCARE APPLICATIONS
\subsubsection{Applications} The training stage has seen the most development of UQ methods, with various approaches used to improve reliability and robustness in medical domains. 

% EDL
Evidential Deep Learning (EDL) has been widely applied across different tasks for its ability to provide meaningful uncertainty estimates while maintaining high performance. 
For instance, \cite{ren2023uncertaintyinformedmutuallearningjoint} used EDL for joint image classification and segmentation in ophthalmology and \cite{yang2023uncertainty} employed it to improve segmentation reliability in general surgery. 
In oncology, \cite{dong2024uncertainty} adopted EDL for prostate cancer grading, while \cite{jeong2024uncertainty} incorporated pixel-wise uncertainty into diffusion models for adversarial colonoscopy image generation.

% Bayesian 
Bayesian methods have also emerged as prominent UQ techniques at the training stage. 
\cite{zhao2022efficient} applied Bayesian techniques for efficient uncertainty estimation in cardiology segmentation tasks, while Bayesian variational inference was leveraged by \cite{adams2023fully} for super shape prediction in cardiology. 
Additionally, sparse Bayesian networks were proposed by \cite{abboud_sparse_2024}, to efficiently quantify uncertainty in skin cancer classification and chest radiology segmentation.

% Monte Carlo + ensemble
Monte Carlo-based approaches and ensemble methods are also commonly used for UQ during training. 
Monte Carlo (MC) Dropout was utilized by \cite{aljuhani_uncertainty_2022} for histological image classification in oncology. 
Ensemble methods were explored by \cite{kazemi2022deue} on video frames to address ejection fraction regression in cardiology.
Similarly, \cite{wu2022nonlinear} combined Bayesian and ensemble techniques to predict remaining surgery duration in ophthalmology using surgical video frames, producing a prediction and uncertainty estimation in the same inference run. 
In another vision application, \cite{zhao2024lost} combined Bayesian methods with ensembles for uncertainty-guided segmentation tasks in cardiology.

% other: RL, AE, prototype, attention module
Beyond these core methods, several innovative techniques have been introduced to address domain- or task-specific challenges.
For example, uncertainty attention modules were employed by \cite{xie2022uncertainty} to handle ambiguous boundary segmentation in cardiology and fetal ultrasound. 
Uncertainty-weighted class activation maps were leveraged by \cite{fu2025cam} for weakly-supervised segmentation in neonatal medicine. 
Epistemic and aleatoric uncertainties were explicitly modeled by \cite{xiang2022fussnet}, who employed these techniques for supervised and unsupervised learning in segmentation tasks within pancreatology and cardiology. 
\cite{larrazabal_maximum_2023} proposed regularization techniques, such as the use of maximum entropy calibration to improve segmentation reliability in radiology and cardiology.
Moreover, contrastive learning and latent space comparisons were developed by \cite{judge2022crisp}, enabling robust uncertainty estimation across multiple imaging datasets.

Entropy-based methods are also frequently employed for UQ.
\cite{sharma_confidence_2024} introduced entropy-driven self-distillation learning to enhance classification performance in ophthalmology, oncology, and skin cancer applications. 
Gaussian probability distributions were leveraged by \cite{judge2023asymmetric}, addressing challenges in cardiology and chest X-ray image segmentation, while \cite{li_dynamic_2023} proposed a Dirichlet distribution classifier for curriculum learning in skin cancer and COVID-19 image classification tasks.

Other notable methods include the use of autoencoders by \cite{lennartz_segmentation_2023}, who introduced a segmentation distortion measure to improve uncertainty estimation under domain shifts in neurology imaging. 
Generative adversarial networks (GANs) have also been adapted by \cite{upadhyay2021uncertainty}, who introduced Uncertainty-Guided Progressive GANs (UP-GAN) for image-to-image translation tasks in medical imaging. 
\cite{hung_cross-slice_2024} employed cross-slice attention and evidential critical loss for prostate cancer detection using MRI scans. 
Finally, \cite{browning_uncertainty_2021} used deep reinforcement learning to estimate uncertainty for pathology landmark detection in orthopedics. 
\cite{zou2022tbrats} utilized subjective logic theory to achieve trusted brain tumor segmentation in neurology-oncology.

\subsubsection{Opportunities}

\paragraph{Architecture-Level Approaches.} Recent advancements focus on novel architectures to address specific challenges for UQ.
% \cite{dai_semi-supervised_2023} focused on improving the scalability of BNNs by combining them with variational model ensembling for semisupervised regression to reduce prediction noise and generate more robust pseudo-labels.
\cite{rudner2022tractable} introduced tractable function-space variational inference for BNNs, resulting in improved computational efficiency while maintaining robust uncertainty estimates.
\cite{mao_uasnet_2021} proposed UAS-Net, an architecture that adapts sampling based on uncertainty estimation to help handle noisy data and refine depth estimation accuracy. 
More commonly, existing architectures are modified to adapt UQ methods to specific applications. 
For example, \cite{zhang_uncertainty_2022} integrated transformers with uncertainty modeling using attention mechanisms that adapt based on uncertainty levels in the data.
Similarly, \cite{zhu_robust_2022} designed a depth completion network that incorporates uncertainty into the architectural layers, and \cite{ma_uncertainty-aware_2024} integrated  UQ into the GAN architecture with data augmentation techniques to improve model stability and performance. 
MC dropout has also found widespread use across a variety of applications due to its simplicity and effectiveness.
For example, \cite{tolle_funavg_2024} introduced a federated uncertainty-weighted averaging method, combining Bayesian methods with MC dropout to address diverse label distributions in federated learning systems. 
Similarly, an uncertainty-driven dropout was introduced by \cite{feng2021uag} to enhance the robustness of Graph Neural Networks (GNN) through stochastic techniques.
 
\paragraph{Loss Modification.} Modifying loss functions enables the explicit incorporation of uncertainty into ML models, enhancing their robustness and reliability.
For instance, \cite{warburg_bayesian_2021} introduced a Bayesian triplet loss to generate stochastic embeddings for image retrieval tasks. 
Similarly, \cite{do_semi-supervised_2021} proposed loss modifications for semi-supervised learning that adapt to uncertain labels, effectively reducing overfitting to noisy data. 
In the context of GNNs, \cite{feng2021uag} developed adaptive loss functions that respond dynamically to adversarial perturbations based on uncertainty. 
\cite{caldeira_deeply_2020} investigated how different loss functions, such as Mean Squared Error (MSE) and log-likelihood, influence uncertainty estimation and model robustness in Bayesian frameworks. 

These approaches highlight the growing focus on embedding UQ directly into model design, presenting potential avenues of research that can significantly improve the design and development of robust, tailored UQ methods for model training in healthcare applications. 

%%%%%%%%%%%%%%%%%%%%%%%%%%%%%%%%%%%%%%%%%%%%%%%%

% MODEL EVALUATION
\subsection{Model Evaluation}

The evaluation stage is key for assessing how UQ translates into real-world insights, focusing on model confidence, calibration, and robustness. 
Uncertainty decomposition techniques can also help distinguish between epistemic and aleatoric uncertainty, offering deeper insights into model behavior and guiding clinical improvements at the inference stage.

% PARAGRAPH ON HEALTHCARE APPLICATIONS
\subsubsection{Applications}
% Various methods have been proposed for uncertainty estimation at the model evaluation stage, with many healthcare-specific solutions proposed to reliably estimate uncertainty for clinical applications.

Recent advances have focused on improving predictive uncertainty estimation, Out-Of-Distribution (OOD) detection, and interpretability. 
For instance, \cite{hu2021learning} decomposed prediction error into random and systematic components, proposing a two-step method that estimates target labels and error magnitude, evaluating the method on an MRI reconstruction task.
\cite{teichmann2024towards} introduced a statistical method for OOD detection and for improving the precision of contouring target structures and organs-at-risk, showing that epistemic uncertainty estimation is highly effective for radiotherapy workflows.
\cite{kushibar2022layer} introduced an image-level uncertainty metric to improve uncertainty estimation in segmentation tasks compared to the commonly used pixel-wise metrics such as entropy and variance, validating their method on oncology and cardiology applications. 
To help in medical image understanding, \cite{chen2024modeling} proposed an efficient conformal prediction method along with an uncertainty explanation method to identify the most influential training samples, offering a more interpretable uncertainty estimate for organs and blood imaging datasets.

In orthopedic imaging applications, \cite{wahlstrand_skarstrom_explainable_2024} aligned model uncertainty estimates with intra-reader variability, demonstrating reliability comparable to human annotators, and providing calibrated uncertainty maps to enhance interpretability in vertebral fracture assessment. 
\cite{yang2022uncertainty} considered the information present in annotations introducing a multi-confidence mask, to predict regions with varying uncertainty levels in lung nodule segmentation, suggesting that regions causing segmentation uncertainty are not random but are related to disagreements in radiologist annotations. 
Similarly, \cite{konuk2024framework} argued that current uncertainty evaluation metrics fall short in clinical contexts, and proposed an evaluation framework to inform joint human-AI systems.
To tackle overconfident predictions, \cite{popordanoska2021relationship} investigated the relationship between calibrated predictions and volume estimation in medical image segmentation, validating their findings on glioma and ischemic stroke lesion volume estimation.
For example, to address the lack of UQ methods that are adapted to precision medicine, \cite{improving_image_based_precision_medicine} used Bayesian deep learning to assess model uncertainty in MRI scans for multiple sclerosis, correlating predictive uncertainty with treatment options to enhance clinical decision-making.
Finally, in an effort to improve the evaluation of UQ methods on real-world applications, \cite{band2022benchmarking} built an open-source benchmark for diabetic retinopathy detection tasks. 
Their benchmark uses a set of task-specific reliability and performance metrics to evaluate Bayesian methods on safety-critical scenarios, reflecting the complexities of real-world clinical data.

In EHR-based applications, \citet{ehr_1} estimated heterogeneous treatment effects using a Bayesian Gaussian-process-based partially linear model, enabling fine-grained UQ in observational data. 
\citet{ehr_2} integrated variational dropout and deep ensembles to enhance both calibration and counterfactual decision-making. 
To capture temporal dynamics, \citet{ehr_4} modeled patient trajectories through a Bayesian neural controlled differential equation framework, quantifying both model and outcome uncertainty. 
Additionally, \citet{ehr_6} employed Gaussian random fuzzy numbers within an evidential regression model to simultaneously estimate epistemic and aleatoric uncertainties for time-to-event prediction.

\subsubsection{Opportunities}

\paragraph{Post-hoc Calibration.} 
Methods such as test-time augmentation are used to improve model calibration by generating diverse inputs for model evaluation to improve generalization \citep{hekler_test_2023}. 
Dirichlet-based models also help adjust the model's output by recalibrating probabilities \citep{shen_post-hoc_2023, kopetzki2021evaluating}. 
Additionally, \cite{li_uncertainty_2022-1} proposed a response-scaling method of the input to improve the numerical stability of UQ methods, and enhancing the overall reliability of the predictions.

\paragraph{Bayesian, Ensemble and Probabilistic Methods.} 
Monte Carlo methods have been used extensively across different domains to provide uncertainty estimates by simulating dropout during inference \citep{zheng_continual_2021, bethell_robust_2024, oberdiek2022uqgan, wagh2022evaluating}. 
Moreover, \cite{yao_stacking_2020} used Bayesian stacking during inference to construct a weighted average of posterior distributions. 
Additionally, \cite{dai_semi-supervised_2023} designed loss functions that integrate uncertainty consistency with Bayesian ensemble methods, enabling robust pseudo-labeling and improved performance in settings with limited supervision.

\subsection{Comparative Evaluation of UQ Methods in Clinical Applications}
\label{sec:comparative_analysis}
In this section, we compare and highlight seven widely used UQ techniques, discussing strengths, limitations, and suitability for healthcare applications based on key criteria such as predictive performance, calibration, scalability, robustness, and computational cost.

BNNs model both epistemic and aleatoric uncertainty but are computationally intensive and difficult to scale \citep{antoran_getting_2021, morales-alvarez_activation-level_2021}. 
They have been applied to medical image classification, disease progression modeling, and decision support \citep{adams2023fully, zhao_lost_2024}. 
GPs provide well-calibrated estimates but do not scale well to large datasets. 
In healthcare, they are used for disease progression prediction, time-series forecasting, and biomarker discovery \citep{wang_uncertainty_2020, peluso_deep_2024}.
Ensemble methods quantify uncertainty via inter-model variability, improving robustness in image segmentation, classification, and anomaly detection, though they lack a principled Bayesian formulation and are computationally costly \citep{dusenberry_analyzing_2020, vranken_uncertainty_2021}. 
Evidential Deep Learning captures both types of uncertainty in a single pass and is used in autonomous diagnostics and decision support, but it may yield overconfident predictions without proper regularization \citep{shi_evidential_2024, hung_cross-slice_2024}.
Conformal Prediction offers distribution-free confidence intervals based on past errors and is applied in risk assessment and predictive modeling, although its guarantees rely on the assumption that past data distributions hold \citep{dutta_estimating_2023, stutts_mutual_2024}.
Bayesian Deep Ensembles improve calibration and epistemic uncertainty estimation over standard ensembles, but their high computational cost limits real-time use \citep{wilson2020bayesian}. 
Monte Carlo Dropout approximates Bayesian inference with lower overhead and is widely used in imaging and predictive modeling, though its estimates depend on the choice of dropout rate and may not fully capture uncertainty \citep{abdar_uncertainty_2021, bethell_robust_2024}.
While many of these methods demonstrate strong empirical performance, their practical adoption in healthcare requires careful consideration of trade-offs between computational cost, scalability, interpretability, and robustness to ensure reliable clinical deployment.

Our systematic analysis reveals that strategically integrating UQ into ML pipelines in healthcare, whether during data processing, model training, or evaluation, has tremendous potential to enhance clinical workflow efficiency and ensures that uncertainty is addressed at the right stage. 
% This can help further advance UQ methodologies in healthcare and also strengthen the robustness and reliability of ML models for integral patient care. 

\section{Open Challenges and Future Research}
\label{sec:future}
% Subsections storyline
% 1. Diversity in clinical datasets: Keep early — data is foundational
% 2. Uncertainty stratification: Understanding uncertainty sources logically follows
% 3. Unified framework across pipeline: Need to systematically manage uncertainty
% 4. Tailored methods for healthcare: Call for healthcare-specific innovation
% 5. Interpretability: Make UQ actionable for clinicians
% 6. Fairness and bias mitigation: Social analysis of UQ
% 7. Deployment considerations: Transition from research to real-world use
% 8. Safety and risk management: Critical issues during deployment
% 9. Standardized evaluation: Need for reproducible evaluation methods
% 10. Regulatory incorporation: Institutionalize best practices
% 11. Roadmap forward: Final synthesis
\subsection*{Expanding Clinical Dataset Diversity}
Despite a growing use of UQ methods in healthcare, most applications remain concentrated on medical imaging, particularly MRI \citep{acdc}, CT \citep{KiTS19}, and X-rays \citep{vindrcxr}, limiting generalizability across other healthcare domains. 
While relevant imaging applications such as skin cancer detection \citep{ren_skincon_2024} and brain tumor segmentation \citep{fuchs2021practical} are well-studied, research into other modalities, such as sensor data and ECG, remains limited.
Decentralized learning approaches, including federated, swarm, and split learning, can offer privacy-preserving solutions for efficient datasharing in healthcare AI \citep{antunes2022federated}. 
Key directions include developing standardized UQ frameworks for decentralized settings, improving calibration across heterogeneous non-iid datasets, and designing lightweight UQ methods to mitigate communication and computational overhead \citep{antunes2022federated, nguyen2022federated}. 
While federated learning applications exist in healthcare \citep{nguyen2022federated}, further research is needed to fully extend UQ into decentralized frameworks.

Future work should prioritize diversifying datasets and clinical tasks to broaden UQ applicability \citep{loftus_uncertainty-aware_2022} and explore multimodal UQ methods \citep{dutta_estimating_2023, jung2024beyond} for more robust and realistic clinical models.

\subsection*{Analyzing Sources of Uncertainty} 
% CHALLENGES
Most research focuses on quantifying predictive uncertainty \citep{lakshminarayanan_simple_2016}, but few studies address understanding its origins, an essential step for trustworthy AI systems.
% FUTURE RESEARCH
Efforts should focus on identifying whether uncertainty stems from data noise \citep{alizadehsani_handling_2024}, model specification issues \citep{do_semi-supervised_2021}, or training data limitations \citep{huang_uncertainty-aware_2022, macdonald2023generalising}. 
Understanding these sources can guide better clinical decision support and model design.

\subsection*{Building a Unified Framework for UQ Across the ML Pipeline} 
% CHALLENGES
Our analysis shows that UQ is typically applied in isolation at different ML stages, particularly during training \citep{abboud_sparse_2024, aljuhani_uncertainty_2022}, with little integration across preprocessing \citep{angelopoulos2022image} and evaluation \citep{hu_learning_2021}.
This fragmentation leads to uncertainty propagation and compounded errors \citep{valdenegro2024unified}.
A unified pipeline-oriented approach would enable systematic uncertainty management, categorizing and evaluating methods at each ML stage, identifying gaps where specific uncertainties remain unaddressed \citep{gruber_sources_2023, jurgens2024epistemic} and guiding the development of holistic solutions.

\subsection*{Developing Tailored UQ Methods for Healthcare}
Current UQ studies often focus on empirical gains without advancing theoretical foundations or understanding limitations in clinical contexts \citep{improving_image_based_precision_medicine, teichmann2024towards}. 
A balanced approach addressing both theory and application is critical. 
Further refinement of popular methods such as deep ensembles \citep{abdollahi2021deep, gu2021predicting}, MC dropout \citep{bethell_robust_2024}, and BNNs \citep{herzog2020bayesian} is needed, alongside development of novel adaptations for underexplored healthcare domains.

\subsection*{Enhancing Interpretability in UQ for Healthcare}
Interpretability of UQ in clinical settings remains underdeveloped. 
Although uncertainty and noise are often intertwined, distinguishing true uncertainty (i.e., aleatoric, epistemic) from noise due to data variations (e.g., incorrect measurements, missing labels) or model training (e.g., parameter selection) is crucial for actionable insights \citep{xiang_fussnet_2022, zhang_informative_2023}. 
Yet, no consensus exists on clinically meaningful UQ metrics, and many approaches lack clinician input. 
Research still heavily focuses on training-stage uncertainty, with limited exploration at inference and deployment, where clinical decisions occur \citep{angelopoulos2022image, kushibar2022layer, konuk2024framework, leibig2022combining}. 
Stronger interdisciplinary collaboration between AI researchers and clinicians is needed, to ensure UQ methods deliver clinically actionable information.

\subsection*{Mitigating Fairness and Bias Challenges}
Bias in AI-driven healthcare arises from multiple sources, including data (e.g., underrepresentation of certain populations), algorithmic (e.g., modeling choices amplifying disparities), and selection biases (e.g., systematic exclusions in data collection) \citep{tripepi2010selection, gianfrancesco2018potential, chen2024unmasking}. 
% UQ methods can provide a framework to assess and mitigate these biases by enabling calibrated decision-making and improving model interpretability with different principles. 
Evaluating uncertainty across demographic subgroups can reveal discrepancies in model confidence, identifying populations for which predictions are systemically unreliable \citep{bozkurt2020reporting}, therefore adjusting decision thresholds to ensure consistent predictive performance across diverse patient cohorts and clinical settings \citep{10.1145/3716317}.
While efforts to mitigate bias are gaining traction, current UQ methods often lack systematic evaluations across diverse demographic groups. 
Future research should prioritize stratified uncertainty analyses to ensure models maintain consistent reliability across age, sex, ethnicity, and disease subpopulations. 

\subsection*{Ensuring Safety and Risk Management in Clinical Applications}
The risk profile of healthcare applications dictates the required level of UQ rigor, reliability and interpretability, since decisions can have life-altering consequences for patients \citep{huang_review_2024}. 
In wellness applications (e.g., fitness trackers, general health monitoring), UQ can improve transparency as errors have lower stakes despite the potential of misleading information for the user \citep{Zhou_Review}. 
In safety-critical domains such as radiology or surgery, UQ must provide high reliability and clinical guarantees \citep{khalighi2024artificial}. 
Regulatory frameworks should differentiate between wellness tools and safety-critical AI, enforcing stronger UQ integration where patient safety is more critical. 
Adaptive methods such as conformal prediction and deep ensembles \citep{zhou2024safety, thompson2025early} can support real-time clinical decision-making, ensuring uncertainty aligns with dynamic clinical contexts.
In addition, thresholds for uncertainty alarms should be adapted based on clinical application and risk level, with regulatory standards working towards enforcing safeguards to ensure AI reliability in safety-critical settings.

\subsection*{Establishing a Standardized Framework for Evaluation}
A structured framework to evaluate UQ methods in healthcare is essential, given the diversity of clinical tasks, datasets, and models \citep{eval_uncertainty_1, eval_uncertainty_3, eval_uncertainty_4}.
To address challenges such as inconsistent evaluation metrics \citep{eval_uncertainty_3}, limited generalizability \citep{eval_uncertainty_1}, and lack of OOD benchmarking \citep{eval_uncertainty_1}, we propose a framework comprising four core components.
The \textbf{first component} focuses on standardized performance-based metrics for uncertainty-aware predictions.
Although many studies assess UQ indirectly via task performance (e.g., classification, segmentation) \citep{eval_uncertainty_1, eval_uncertainty_7}, evaluation protocols should consistently report traditional metrics such as accuracy, precision-recall, F1-score, and AUC.
The \textbf{second component} emphasizes human-machine interaction metrics, particularly selective prediction.
UQ enables selective deferral of uncertain cases to human experts, improving clinical decision-making \citep{eval_uncertainty_1}.
Comparative evaluations should assess deferral strategies and their impact, especially in high-risk scenarios.
The \textbf{third component} addresses calibration analysis to ensure the clinical reliability of uncertainty estimates.
Well-calibrated predictions are critical to prevent misleading outputs \citep{eval_uncertainty_1, eval_uncertainty_4, eval_uncertainty_5}.
Metrics such as Expected Calibration Error (ECE) and Brier scores should be systematically reported across clinical tasks, yet are often overlooked \citep{eval_uncertainty_5, benchmarking_biosignals}.
The \textbf{fourth component} involves OOD detection to assess model robustness under distributional shifts.
Given the frequent domain shifts in healthcare applications, evaluation should explicitly test models’ ability to distinguish in-distribution from OOD samples \citep{eval_uncertainty_1, benchmarking_biosignals}.
Methods such as controlled data perturbations \citep{benchmarking_biosignals} can facilitate systematic OOD benchmarking on clinical datasets.
While domain-specific adaptations may be necessary, adopting a standardized evaluation framework would substantially improve reproducibility, robustness, and trustworthiness of UQ methods, particularly for long-term clinical deployment.

\subsection*{Integrating UQ into Regulatory Frameworks}
Current AI healthcare regulations emphasize transparency, reliability, risk management, and patient safety principles, although they rarely explicitly mention UQ \citep{schmidt2024mapping}. 
The FDA, Health Canada, and the UK’s MHRA have issued ``\textit{Transparency for Machine Learning-Enabled Medical Devices: Guiding Principles}," emphasizing interpretability, reliability, and adaptability, which are core principles underlying the objectives of UQ \citep{us2024transparency}. 
Additionally, the World Health Organization (WHO) has highlighted the importance of regulatory oversight in AI-driven health applications, emphasizing transparency and risk mitigation as central elements \citep{tsaneva2025decoding}. 
In particular, UQ can optimize workflows by enabling uncertainty-adapted triage and optimization risk stratification, ensuring ambiguous cases receive additional scrutiny before critical decisions are made. 
Experts have also argued that AI regulations should explicitly require uncertainty-aware metrics to ensure the safe deployment of AI models, alongside task-specific continuous monitoring protocols \citep{chua2023tackling}.
In the future, holistic regulatory frameworks should include benchmarks for calibration, OOD detection, and selective deferral.

\subsection*{Addressing Deployment Challenges}
While UQ in healthcare AI has been extensively studied from a theoretical perspective, its real-world clinical deployment remains limited. 
Most current research emphasizes potential benefits and challenges but offers limited practical implementation strategies.

Several key barriers must be addressed to translate UQ advances into clinical practice. 
First, real-time clinical decision support requires UQ methods that are computationally efficient and scalable, capabilities that many existing techniques lack \citep{verma2021implementing}.
Second, healthcare data quality and accessibility issues, such as noise, incompleteness, and privacy restrictions, complicate the development of reliable uncertainty estimates \citep{zhang2022shifting}.
Decentralized learning approaches offer promising solutions by enabling robust training without sharing raw data across institutions \citep{yuan2024toward}.
Third, the lack of open science practices, including limited code and model sharing (Figure \ref{fig:figure_1}c), hinders transparency, reproducibility, and comparative evaluation.
Finally, the absence of standardized UQ evaluation frameworks in healthcare leads to inconsistencies across studies, complicating clinical translation. 
Addressing these barriers requires closer collaboration between ML researchers and clinicians to align UQ with real-world clinical needs and operational constraints.

Focused efforts on computational efficiency, data robustness, open benchmarking, and standardized evaluation can significantly advance the integration of UQ into clinical AI workflows, ultimately improving trust, reliability, and patient outcomes.
Addressing these challenges highlights the need for a structured roadmap to guide future research and facilitate the practical deployment of uncertainty-aware AI systems in clinical settings.

\subsection*{Defining a Roadmap for Future Research}
Building on the identified challenges and regulatory considerations, the shift from theoretical UQ research development to real-world clinical deployment requires strategic advancements in evaluation, interpretability, communication, and integration into decision-support systems. 
We outline four actionable research directions to facilitate this transition:

\begin{enumerate}
    \item \textbf{Standardizing UQ Evaluation Metrics:} The lack of consistent evaluation metrics across clinical tasks hinders comparability. Reporting guidelines should mandate standardized assessment of uncertainty calibration, coverage error, out-of-distribution OOD detection, ensuring that uncertainty is evaluated alongside performance metrics in a structured manner.
    \item \textbf{Contextualizing UQ with Task-Specific Safety Thresholds:} UQ must be aligned with the safety-critical nature of specific clinical applications. Regulatory bodies and clinical experts should define impact-sensitive uncertainty thresholds to ensure that AI models meet appropriate patient safety standards.
    \item \textbf{Enhancing Interpretability and Clinical Trust in UQ:} Uncertainty estimates must be both clinically meaningful and interpretable. Developing intuitive visualizations and fostering close collaboration with clinicians on UQ interpretation can bridge the gap between AI model outputs and real-world clinical decision-making.
    \item \textbf{Integrating UQ into Assistive AI and Decision Support:} AI systems should leverage UQ to highlight high-uncertainty cases, allowing clinicians to exercise greater caution where needed. Future efforts should prioritize interpretable, impact-sensitive UQ methods developed collaboratively with clinicians and patients to ensure practical utility.
\end{enumerate}

\section{Conclusion}
\label{sec:conclusion}
In this survey, we reviewed and synthesized recent advancements in UQ methods for healthcare, providing a comprehensive analysis of their application across the ML pipeline. 
We discussed popular methodologies, key clinical domains, relevant medical datasets, and outlined current challenges alongside promising future research directions. 
We emphasize the need for an integrated and systematic approach to incorporate UQ across all stages of model development for clinical applications. We encourage researchers to evaluate proposed algorithms across diverse medical datasets, integrate UQ techniques throughout the ML pipeline, and conduct detailed analyses of uncertainty sources to more effectively mitigate them. 
Our recommendations aim to bridge existing research gaps and guide future work in UQML4H, ultimately supporting the development of trustworthy, reliable, and clinically meaningful ML systems.

\clearpage

\section*{Acknowledgements}
This work was supported by the NYUAD Center for Artificial Intelligence and Robotics, funded by Tamkeen under the NYUAD Research Institute Award CG010.
\bibliography{references}

\clearpage
\onecolumn
\appendix

\section{Foundations of Uncertainty Quantification in Machine Learning}
\label{sec:foundations_uq}
\setcounter{table}{0}
\renewcommand{\thetable}{A\arabic{table}}

\setcounter{figure}{0}
\renewcommand{\thefigure}{A\arabic{figure}}

\begin{figure*}[h!]
    \centering
    \includegraphics[width=\textwidth]
    {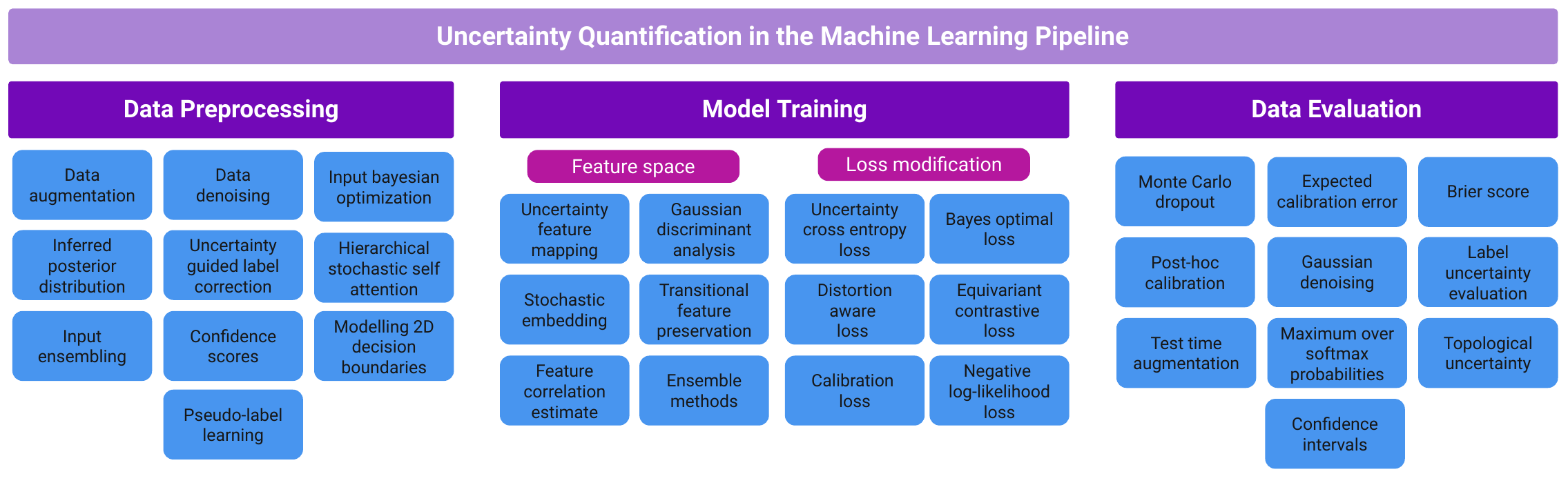}
    \caption{\textbf{Uncertainty quantification across the ML pipeline}. 
    Key UQ methods from different domains applied at each stage: data processing, model training, and evaluation.} 
    \label{fig:figure_2}
\end{figure*}

This appendix provides a brief overview of key UQ methods developed across different machine learning applications. 
We do not aim at a comprehensive review, as previous work already covered the theoretical foundations of UQ in great depth \citep{gruber_sources_2023, liu2018review, seoni_application_2023}. 
Instead, we focus on presenting key methods and trade-offs relevant for readers interested in applying UQ techniques, particularly in healthcare contexts as described in Section \ref{sec:ml_pipeline}.
Figure \ref{fig:figure_2} summarizes the common UQ approaches mapped to each stage of the machine learning pipeline.

\subsection{Probabilistic Methods}
% Introduction of the type of methods
These approaches represent uncertainty using probability distributions and statistical models, providing a mathematical framework for modeling variability and randomness in the data and model architecture.

\paragraph{Bayesian Neural Networks (BNNs).}
BNNs learn distributions over network weights rather than relying on fixed point estimates, providing a principled Bayesian framework for modeling uncertainty.
Training is computationally expensive due to intractable posterior inference, often requiring approximations that may affect calibration \citep{blundell_weight_2015, gal_dropout_2015}.
Inference can also be slow depending on the approximation method used.
BNNs have been applied to clinical tasks such as outcome prediction and survival analysis, where quantifying model confidence is critical \citep{wang2020review, herzog2020bayesian}.

\paragraph{Gaussian Processes (GPs).}
GPs are non-parametric models that define distributions over functions fitting the data, offering strong uncertainty estimates, especially for small datasets.
They provide exact Bayesian inference but scale poorly ($O(n^{3})$ complexity), limiting applicability to large-scale problems \citep{dietterich2024uncertainty, liu2018review}.
GPs have been employed in modeling disease trajectories and personalized medicine applications \citep{futoma2018gaussian, puri2022forecasting}.

\paragraph{Ensemble Methods.}
Ensemble techniques improve predictive performance and uncertainty estimation by aggregating outputs from multiple models \citep{dietterich2000ensemble}.
Approaches include deep ensembles \citep{lakshminarayanan_simple_2016}, bagging \citep{rokach2010ensemble}, and snapshot ensembles \citep{zhou2012ensemble}.
Ensembles require training multiple models independently, resulting in high memory and training costs, and inference overhead from multiple forward passes \citep{dietterich2000ensemble}.
They have demonstrated success in diagnostic tasks such as pneumonia detection from chest X-rays and sepsis prediction \citep{shilo2020applications, valk2021uncertainty}.

\subsection{Non-Probabilistic Methods}

These methods quantify uncertainty without relying on explicit probability distributions, often using bounded sets or evidence-based frameworks.

\paragraph{Evidential Deep Learning (EDL).}
EDL incorporates uncertainty estimation into the learning process by modeling evidence through Dirichlet distributions.
EDL methods typically avoid sampling-based inference, leading to faster training and single-pass inference while maintaining theoretical rigor \citep{sensoy2018evidential}.
They have been applied to disease classification tasks, enabling models to flag uncertain predictions for clinical review and improving AI-driven diagnostic safety \citep{deng2023uncertainty}.

\paragraph{Fuzzy Logic.}
Fuzzy logic captures uncertainty by modeling partial membership across multiple classes, useful for handling imprecise or vague information.
Training complexity varies with rule complexity but is generally lower than probabilistic approaches; inference is fast but lacks probabilistic confidence estimates \citep{zadeh1988fuzzy, kosko1993fuzzy}.
In healthcare, fuzzy logic has been applied to clinical decision systems where test results or symptom categories are inherently ambiguous \citep{nguyen2015classification, gursel2016healthcare}.

%%%%%%%%%%%%%%%%%%%%%%%%%%%%%%%%%%%%%%%%%%%%%%%%%%%%%%%%%%%%%%%%%%%%%%%%%%%%%%%%%%%

\subsection{Hybrid Methods}

% Introduction of the type of methods
These methods are combinations of different probabilistic and non-probabilistic approaches for UQ that leverage the strengths of each framework. 

\paragraph{Bayesian Deep Ensembles.} Combine the strengths of ensemble learning and Bayesian learning, leveraging the diversity of multiple indepoendently trained models with Bayesian principles while incorporating priors or randomized initialization \citep{wild2023rigorous, abulawi2024bayesian}. 
They have been used in the diagnosis of chronic diseases \citep{abdollahi2021deep} and prediction of medication adherence \citep{gu2021predicting}.

\paragraph{Conformal Prediction.} By constructing prediction sets or intervals that contain the true label with a user-specified confidence level, conformal predictors offer a formal measure of uncertainty with guaranteed coverage probabilities \citep{angelopoulos2021gentle}. 
This approach has been employed to enhance the confidence in predictions for skin lesions \citep{lu2021fair} and genomics \citep{papangelou2024reliable}.

%%%%%%%%%%%%%%%%%%%%%%%%%%%%%%%%%%%%%%%%%%%%%%%%%%%%%%%%%%%%%%%%%%%%%%%%%%%%%%%%%%%%

\subsection{Key Domains} 

In reviewing SOTA UQ methods, several key application domains emerge and can be summarized as follows.  

\paragraph{Mathematical Foundations.} This represents the backbone of the UQ field, focusing on developing rigorous theoretical frameworks, probabilistic models, and algorithmic proofs to enhance uncertainty modeling, model calibration, and learning stability \citep{wang_energy-based_2021, pei_transformer_2022, ghosh_improving_2023, arora_leveraging_2024}.
Innovative approaches include normalizing flows, function space priors, and infinite-width neural networks to improve both epistemic and aleatoric uncertainty estimation \citep{bae_estimating_2021, adlam_exploring_2021, berry_normalizing_2023, schnaus2023learning}. 
Studies aiming to produce more reliable and interpretable predictions focus on methods such as conformal prediction, test time augmentation, mutual information, temperature scaling, and ensemble learning \citep{kuleshov2022calibrated, hekler_test_2023, li_calibrating_2023, wang_learning_2023, stutts_mutual_2024}. 

\paragraph{Optimization.} Closely linked to mathematical foundations, optimization is a widely studied domain to refine model performance, calibration, and generalization. 
Techniques such as gradient-based optimization, regularization strategies, and novel loss functions are employed to mitigate overfitting and calibrate predictions during training \citep{heiss2102nomu, xia2021sample, dai_semi-supervised_2023, daheim2023model}.

\paragraph{Computer Vision (CV).} CV is a major application domain where UQ methods are applied to critical tasks such as crowd counting, image segmentation, depth estimation, image denoising, multi-view stereo, and medical imaging \citep{qu2021bayesian, mao2021uasnet, manor2023posterior, li_calibrating_2023, kahl2024values,wang2022uncertainty}. 
Other applications focus on identifying ambiguous or out-of-distribution data, object tracking, image-to-image regression, vision-matting and facial expression recognition \citep{Hu2020Unsupervised, zhang2021relative, nussbaum2022structuring, angelopoulos2022image, zhang2022uast, wu2023dugmatting}. 

\paragraph{Natural Language Processing (NLP).} 
% Despite the growing prominence of large language models (LLMs), the development of uncertainty quantification methods in NLP remains limited.w
Relevant methods in NLP domain focus on uncertainty at the level of token-level prediction, text generation, dialogue retrieval, code generation, and LLM fine-tuning, addressing challenges related to confidence estimation and calibration both during data processing and model training \citep{malinin2020uncertainty, hou2023decomposing, xiong2023can, johnson2023ru, gupta2024language, lee2024improving, liuuncertainty}.

\paragraph{Reinforcement Learning (RL).} Primarily used to optimize the exploration-exploitation trade-off, improve policy robustness, enable multitasking in offline RL, and enhance Q-learning in uncertain environments \citep{wu2021uncertainty, liu2022uncertainty,xie2022robust, bai_pessimistic_2024}. 
Other methods include Bayesian RL, meta-learning, and UQ exploration, which improve the reliability of adaptive learning in dynamic environments \citep{li2021mural, gong_adaptive_2023, zhang2024uncertainty}. 

\paragraph{Graph Neural Networks (GNNs).} GNNs, known for modeling complex relational data, are explored for UQ in applications such as molecular modeling, adversarial robustness and social network analysis \citep{shanthamallu_uncertainty-matching_2021, feng2021uag, yu2023uncertainty, wollschlager2023uncertainty, trivedi2024accurate}.

\paragraph{Multimodal Learning.} 
The development of UQ methods in multimodal learning has been limited by the lack of high-quality, large-scale multimodal datasets, resulting in being constrained to highly specialized applications such as text-to-image person reidentification, sensing in soft robotics systems, malware detection and traffic trajectory planning \citep{brown_uncertainty_2020, ding_predictive_2021, zhao2024unifying, lafage_hierarchical_2024}.
In healthcare, current studies include depression and stress detection, sentiment analysis, mortality prediction, clinical imaging segmentation, and mRNA classification \citep{foltyn2021towards, han2022multimodal, ahmed_taking_2023, bezirganyan2023data, huang_deep_2023}.

\paragraph{Emerging Fields.} 
The application of UQ remains limited in several research areas, including: (1) Generative adversarial networks (GANs) in addressing vulnerabilities to adversarial attacks and enhancing resilience against uncertainty manipulation \citep{hu_multidimensional_2021, galil2021disrupting, schweighofer2023quantification}, (2) federated learning and (3) contrastive learning, being used independently in handling noisy, heterogeneous data while ensuring data privacy and model generalizability \citep{plassier2023conformal, kotelevskii_dirichlet-based_2024, wang2024bridging}. 
Additionally, evidential deep learning has gained increasing attention in recent years for its potential integration into clinical decision support systems to enhance medical diagnostics and risk assessment \citep{deng2023uncertainty, ashfaq2023deed, li2024hyper, jurgens2024epistemic, liu2024weakly}. 
These research areas reflect a dynamic, interdisciplinary effort to develop safe and robust ML models. 
Given the importance of UQ in high-stakes decision-making, we focus on its development and impact in the healthcare domain across each stage of the ML pipeline.

\clearpage
\section{Medical Datasets for Uncertainty Quantification in Healthcare}  
\label{app:medical_datasets}  
\setcounter{table}{0}
\renewcommand{\thetable}{B\arabic{table}}

\setcounter{figure}{0}
\renewcommand{\thefigure}{B\arabic{figure}}
A wide range of medical datasets has been developed to support machine learning research across diverse clinical tasks and conditions. 
These datasets serve as critical benchmarks for training, validation, and evaluation, enabling the development and assessment of uncertainty quantification models in healthcare.  
Table~\ref{table:table_datasets} presents a structured overview of open-access datasets, categorized by medical domain, along with their frequency of use across the reviewed studies. 
A key observation from our analysis is the strong reliance on standardized clinical datasets, largely driven by the practical constraints of medical data accessibility, suggesting that dataset availability often outweighs theoretical considerations in shaping UQ research directions. 
The table also highlights the primary clinical task associated with each dataset, offering a comprehensive reference for researchers integrating UQ methods into medical applications.

\paragraph{Private Datasets.}  
In addition to public datasets, many studies leverage institution-specific private datasets, particularly for imaging-based applications and rare disease research. 
Notable examples include endoscopic submucosal dissection procedures, knee MRI for musculoskeletal analysis \citep{browning_uncertainty_2021}, MRI-to-PET translation \citep{upadhyay2021uncertainty}, fetal brain MRI for neurodevelopmental assessment \citep{fu2025cam}, sleep pattern monitoring \citep{kang_statistical_2021}, and specialized cardiology \citep{adams2023fully} and oncology datasets, such as ovarian and prostate cancer imaging \citep{konuk2024framework, dong2024uncertainty}.
Although not openly accessible, these datasets provide valuable insights into specialized clinical domains, where UQ methods contribute to enhancing diagnostic confidence and decision support.

{\footnotesize
\begin{longtable}{
>{\raggedright\arraybackslash}p{7.5cm}
>{\raggedright\arraybackslash}p{6cm}
>{\centering\arraybackslash}p{2cm}
}

\caption{Summary of Open-Access Healthcare Datasets for Uncertainty Quantification Research} 
\label{table:table_datasets}  \\

\toprule
    % \multicolumn{1}{c}{\multirow{2}{*}{\centering\textbf{Dataset}}} & 
    % \multicolumn{1}{c}{\multirow{2}{*}{\centering\textbf{Clinical Task}}} & 
    % \multicolumn{1}{c}\textbf{No. of Papers $^*$}
    \multicolumn{1}{c}{\textbf{Dataset}} & 
    \multicolumn{1}{c}{\textbf{Clinical Task}} & 
    \multicolumn{1}{c}{\textbf{No. of Papers$^*$}} \\
    \endfirsthead

\toprule
    \multicolumn{1}{c}{\textbf{Dataset}} & 
    \multicolumn{1}{c}{\textbf{Clinical Task}} & 
    \multicolumn{1}{c}{\textbf{No. of Papers$^*$}} \\
    \bottomrule
    \vspace{0.5mm}
    \endhead

    \midrule
    \multicolumn{3}{c}{\textbf{Multi-domain}} \\
    \midrule

    MIMIC-III \citep{mimiciii} & Electronic health records & 1\\[3pt] 
    eICU \citep{eicu} & Multi-center critical care database & 1 \\[3pt] 
    FastMRI \citep{fastmri}  &  Knee, brain, prostate, breast classification & 2\\[3pt] 
    MedMNIST \citep{medmnist}  &  Biomedical classification \& segmentation & 1\\[3pt] 
    CPRD \citep{cprd} & Electronic health records  & 1 \\[3pt] 
    TOP \citep{topclinicaltrial} & Clinical trial outcome prediction & 1 \\[3pt] 
    BioVid \citep{biovid} & Pain assessment & 1 \\[3pt] 
    PAMAP2 \citep{PAMAP2} & Activity monitoring & 1 \\[3pt] 
    USC Alcohol Concentration \citep{alcoholconcentration} & Blood alcohol concentration estimation & 1 \\[3pt]
    MIMIC-IV \citep{mimic_iv} & Time-to-Event Prediction, CATE Estimation & 2 \\[3pt]

    \midrule
    \multicolumn{3}{c}{\textbf{Microscopy}} \\
    \midrule

    TEMCA2 \citep{temca2} & Electron microscopy of adult fly brain & 1\\[3pt] 
    BSCCM \citep{bsccm}  &  Single white cell microscopy & 1\\[3pt] 
    MitoEM \citep{mitoem} & 3D Mitochondria instance segmentation & 1 \\[3pt] 
    Kasthuri++ \citep{kasthuri_lucchi} & Mitochondria segmentation & 1 \\[3pt] 
    Lucchi++ \citep{kasthuri_lucchi} & Mitochondria segmentation & 1 \\[3pt]
    
    \midrule
    \multicolumn{3}{c}{\textbf{Cardiology}} \\
    \midrule
    
    ACDC \citep{acdc}  &  MRI segmentation & 5\\[3pt] 
    HMC-QU \citep{hmcqu} & Myocardial infarction detection in ECG & 2 \\ [3pt] 
    Echonet-Dynamic \citep{echonetdynamic} & Cardiac cycle assessment & 2\\[3pt] 
    ECG5000 \citep{ecg5000} & Congestive heart failure detection & 1 \\[3pt] 
    PhysioNet/CinC Challenge 2020 \citep{physionetcinc} & Cardiac abnormality detection in ECG & 1 \\[3pt] 
    M\&Ms \citep{mnm}  &  Multi-disease cardiac segmentation & 4\\[3pt] 
    CPSC2018 \citep{cpsc2018} & ECG classification & 1 \\[3pt] 
    TMED 2 \citep{tmed2}  &  ECG classification & 2\\[3pt] 
    PTB ECG Database \citep{ptb} & Myocardial infarction detection in ECG & 1 \\[3pt] 
    Atrial Segmentation Challenge \citep{atrialsegmentationchallenge} & Atrial segmentation & 1\\[3pt] 
    CAMUS \citep{camus} & Echocardiographic Image Segmentation & 6\\[3pt]  
    UPL \citep{upl} & Heart MRI segmentation & 1\\[3pt] 
    UCI Heart-Disease \citep{heart_disease_data_set} & Heart disease classification & 1\\[3pt] 
    UCR Time Series Archive \citep{ecg5000} & Heart Disease Detection in ECG & 1 \\[3pt]
    CVSim \citep{cvsim} & Simulating the Dynamics of the Human Cardiovascular System & 1 \\[3pt]

    \midrule
    \multicolumn{3}{c}{\textbf{Gastroenterology}} \\
    \midrule

    CholecSEG8k \citep{cholecseg8k}  &  Cholecystectomy segmentation & 1 \\[3pt] 
    DeepOrgan \citep{deeporgan}  &  Pancreas segmentation & 1\\[3pt] 
    Kvasir-Seg \citep{kvasirseg}  &  Colorectal polyp segmentation & 1\\[3pt] 
    PolypDB \citep{polypdb}  &  Wireless capsule endoscopy detection & 1\\[3pt]

    \midrule
    \multicolumn{3}{c}{\textbf{Neurology}} \\
    \midrule

    IXI \citep{ixi} & Brain MR Images from Healthy Subjects & 1 \\[3pt] 
    ISLES 2018 \citep{isles2018}  &  Ischemic stroke lesion segmentation & 1\\[3pt] 
    WMH Segmentation \citep{challengewmh} & White matter hyperintensities segmentation & 1 \\[3pt] 
    Calgary-Campinas-359 \citep{calgarycampinas359} & Brain segmentation & 1 \\[3pt] 
    BRAVO \citep{bravo} & Evaluating laquinimod in RRMS & 1 \\[3pt] 
    OPERA1 \citep{opera} & Evaluating Treatment Effects in RMS & 1 \\[3pt] 
    DEFINE \citep{define} & Evaluating BG-12 in RMS & 1 \\[3pt]    
    WU-Minn HCP \citep{wuminnhcp} & Characterization of brain connectivity & 1 \\[3pt] 
    HCP Lifespan Studies \citep{lifespan} &  Diffusion MRI images & 1 \\[3pt] 
    INTERGROWTH \citep{intergrowth} & 3D Ultrasound fetal brain volumes & 1 \\[3pt] 
    Prisma \citep{prisma} & MRI Image Enhancement & 1 \\[3pt] 

    IDH-Glioma-MRI \citep{pathology} & IDH Prediction in Brain MRI Images & 1 \\[3pt]

    \midrule
    \multicolumn{3}{c}{\textbf{Oncology}} \\
    \midrule

    BCDR \citep{bcdr} & Benchmarking for Breast Cancer Diagnosis & 1 \\[3pt] 
    BreastPathQ \citep{breastpathq}  &  Breast tumor cellularity assessment & 1\\[3pt] 
    ISIC 2018-2019 \citep{isic18}  &  Skin lesion detection & 7\\[3pt] 
    Breast Histopathology (Kaggle) 
    \citep{breasthistopathologykaggle} & Breast Tumor Histopathology & 1 \\[3pt] 
    KiTS19-21 \citep{KiTS19} & Kidney CT segmentation & 2 \\[3pt]  
    WDBC \citep{wdbc} & Breast cancer classification & 1 \\[3pt]  
    LiTS \citep{livermetastates} & Liver tumor segmentation & 1 \\[3pt] 
    Bone Metastates \citep{bonemetastates} & Bone tumor segmentation & 1 \\[3pt] 
    HAM10000 \citep{ham10000}  &  Skin lesion detection & 2\\[3pt] 
    BraTS 2018-2019 \citep{isic18}  &  Brain tumor segmentation & 4\\[3pt] 
    The Cancer Genome Atlas \citep{tcga} & Different Types of Tumor Detection & 1 \\[3pt] 
    ISPY I \citep{ispy} & Breast Cancer Tumor Segmentation & 1\\[3pt] 
    OrganCMNIST \citep{organcmnist} & Liver Tumor Segmentation & 1\\[3pt] 
    Derm-Skin (DERM) \citep{dermskin} & Skin Cancer Detection & 1 \\[3pt] 
    SkinCon \citep{skincon} & Skin Cancer Detection & 1 \\[3pt] 
    ClinSkin \citep{clinskin} & Skin Cancer Detection & 1 \\[3pt] 
    PAD-UFES-20 \citep{padufes20} & Skin Cancer Detection & 1 \\[3pt] 
    QUBIQ 2021 \citep{qubiq2021} & Skin Cancer Detection & 1\\[3pt] 
    BrainMRI \citep{brainmri} & Brain Tumor Detection & 1 \\[3pt] 
    HECKTOR \citep{hecktor} & Head \& neck tumor segmentation in PET/CT & 1 \\[3pt]
    Tumor Growth Model \cite{simulated_tt} & Time-to-Event Prediction, CATE Estimation & 1 \\[3pt]
    GBSG \cite{gbsg} & Time-to-Event Prediction, CATE Estimation & 1\\[3pt]
    SUPPORT \cite{support} & Time-to-Event Prediction, ITE Estimation & 1\\[3pt]

    \midrule
    \multicolumn{3}{c}{\textbf{Ophthalmology}} \\
    \midrule

    Cataract-101 \citep{cataract101} & Cataract Surgery Videos & 1\\[3pt] 
    EyePACS \citep{eyepacs2015}  &  Diabetic retinopathy detection & 2 \\[3pt] 
    APTOS 2019 \citep{aptos2019}  &  Diabetic retinopathy detection & 2\\[3pt] 
    REFUGE \citep{refuge}  &  Glaucoma assessment& 1 \\[3pt] 
    DPL \citep{dpl} & Fungus Image Segmentation & 1\\[3pt] 
    LAG \citep{lag} & Glaucoma Detection & 1\\[3pt] 
    Diabetic Retinopathy (Kaggle) 
    \citep{kaggle_diabetic_retinopathy} & High-resolution retina images & 1 \\[3pt]

    \midrule
    \multicolumn{3}{c}{\textbf{Pulmonology}} \\
    \midrule

    ChestX-ray8 \citep{chestxray8}  &  Pulmonary disease detection & 2 \\[3pt] 
    TBX11K \citep{tbx11k} & Tuberculosis Diagnosis & 1 \\[3pt]  
    Shenzhen Chest X-ray \citep{shenzen}  &  Pulmonary disease detection & 1 \\[3pt] 
    LIDC-IDRI \citep{lidcidri}  &  Lung nodule detection \& segmentation & 2 \\[3pt] 
    JSRT \citep{jsrt} & Lung Nodules Classification & 2 \\[3pt]  
    RSNA \citep{rsnapneumonia} & Pneumonia Detection & 1 \\[3pt] 
    VinDr-CXR \citep{vindrcxr} & Chest X-Ray Disease Detection & 2 \\[3pt] 
    CXAD \citep{cxad} & Chest X-Ray Disease Detection & 1 \\[3pt] 

    \midrule
    \multicolumn{3}{c}{\textbf{Radiology}} \\
    \midrule
    
    SUPERB \citep{superb1} & Vertebral Fractures Diagnosis & 1 \\[3pt] 
    HC18 \citep{hc18} & Fetal Head Circumference Measurement & 1 \\[3pt] 
    TN-SCUI \citep{tnscui} & Thyroid Segmentation \& Classification & 1\\[3pt]  
    BloodMNIST \citep{bloodmnist} & Disease Classification & 1\\[3pt] 
    Kvasir-SEG \citep{kvasirseg} & Polyp Segmentation & 1\\[3pt] 
    FSM \citep{fsm} & Polyp Segmentation & 1\\[3pt] 
    PICAI \citep{picai} & Prostate Cancer Detection & 1\\[3pt] 
    B-Fract \citep{bfract} & Hairline fracture detection & 1 \\[3pt]  
    Low-Dose CT Images \citep{lowdosect} & Low-dose CT denoising & 1 \\[3pt]

    \bottomrule \\[3pt]
    \multicolumn{3}{l}{* Number of papers in our survey that use the dataset.} \\
\end{longtable}
}  

% \paragraph{Private Datasets}  
% Beyond publicly available datasets, many studies rely on institution-specific private datasets, particularly for imaging-based applications and rare disease cases. 
% Notable examples include endoscopic submucosal dissection procedures, knee MRI for musculoskeletal analysis \citep{browning_uncertainty_2021}, MRI to PET scans for image translation \citep{upadhyay2021uncertainty}, fetal brain MRI for neurodevelopmental assessment \citep{fu2025cam}, sleep pattern data \citep{kang_statistical_2021}, and specialized cardiology \citep{adams2023fully} and cancer imaging datasets such as ovarian and prostate cancer \citep{konuk2024framework, dong2024uncertainty}.
% While these datasets are not openly accessible, they provide valuable insight into specialized medical applications where UQ methods are being implemented, contributing to improved diagnostic confidence and clinical decision-making.

%%%%%%%%%%%%%%%%%%%%%%%%%%%%%%%%%%%%%%%%%%%%%%%%%%%%%%%%%%%%%%%%%%%%%%%%%%%%%%%%%%%%

\clearpage
\section{Healthcare Studies Organized by ML Pipeline Stage}
\label{app:appendix_healthcare_studies}
 \setcounter{table}{0}
\renewcommand{\thetable}{C\arabic{table}}

 \setcounter{figure}{0}
\renewcommand{\thefigure}{C\arabic{figure}}

Table \ref{table:ML_pipeline_healthcare1} summarizes the UQ healthcare studies reviewed in this survey. 
Each study is grouped by medical domain and annotated with the corresponding ML pipeline stage at which UQ methods are applied. 
We also report the specific clinical tasks addressed and the datasets utilized. 
Notably, most studies implement UQ at the model training stage, with a predominant focus on image classification and segmentation tasks.

{\footnotesize
\begin{longtable}{
>{\raggedright\arraybackslash}p{3cm}
>{\centering\arraybackslash}p{0.5cm}
>{\centering\arraybackslash}p{0.5cm}
>{\centering\arraybackslash}p{0.5cm}
>{\centering\arraybackslash}p{3cm}
>{\centering\arraybackslash}p{3cm}
>{\centering\arraybackslash}p{3cm}
}

\caption{Summary of Healthcare Studies Implementing Uncertainty Quantification Methods
\label{table:ML_pipeline_healthcare1}
}  \\

\toprule
        \multicolumn{1}{c}{\textbf{Reference}}
        &  \textbf{Data} & \textbf{Train} & \textbf{Eval} & \multicolumn{1}{c}{\textbf{UQ Method}} & \textbf{Task} & \textbf{Datasets}\\
        \endfirsthead

\toprule
    \textbf{Reference}
        &  \textbf{Data} & \textbf{Train} & \textbf{Eval} & \multicolumn{1}{c}{\textbf{UQ Method}} & \textbf{Task} & \textbf{Datasets}\\
        \bottomrule
        \endhead

    %%%%%%%%%%%%%%% CARDIOLOGY %%%%%%%%%%%%%%% 
    \midrule
    \multicolumn{7}{c}{\textbf{Cardiology}} \\
    \midrule 
    
    \multirow{5}{*}[0.4em]{\citep{gu_reliable_2024}} & \multirow{5}{*}[0.4em]{\checkmark} & \multirow{5}{*}[0.4em]{-} & \multirow{5}{*}[0.4em]{-} & \multirow{5}{*}[0.4em]{Conformal Prediction} & \multirow{5}{*}[0.4em]{Classification} & TMED-2, CIFAR-10-Derived, Private Aortic Stenosis$^*$ \\[3pt]
    
    \midrule 
    \multirow{2}{*}[0.1em]{\citep{oh_uncertainty-aware_2024}} & \multirow{2}{*}[0.1em]{\checkmark} & \multirow{2}{*}[0.1em]{\checkmark} & \multirow{2}{*}[0.1em]{-} & Acoustic Diffusion Method &\multirow{2}{*}[0.1em]{ Segmentation} & Echonet-Dynamic, HMC-QU, CAMUS\\[3pt]
    
    \midrule
    \citep{zhao2022efficient} & -  & \checkmark & -  & Bayesian Learning & Segmentation & ACDC, MnM\\[3pt]

    \midrule 
    \multirow{2}{*}[0.1em]{\citep{zhao2024lost}} & \multirow{2}{*}[0.1em]{-} &\multirow{2}{*}[0.1em]{\checkmark} & \multirow{2}{*}[0.1em]{-} & Bayesian \& Ensemble Learning & \multirow{2}{*}[0.1em]{Segmentation} & \multirow{2}{*}[0.1em]{ACDC} \\[3pt]

    \midrule 
    \multirow{2}{*}[0.1em]{\citep{vaseli2023protoasnet}} & \multirow{2}{*}[0.1em]{-} & \multirow{2}{*}[0.1em]{\checkmark} & \multirow{2}{*}[0.1em]{-} & Prototype Based Models & \multirow{2}{*}[0.1em]{Classification} & \multirow{2}{*}[0.1em]{TMED-2} \\[3pt]

    \midrule 
    \multirow{2}{*}[0.1em]{\citep{lu2023upcol}} & \multirow{2}{*}[0.1em]{-} & \multirow{2}{*}[0.1em]{\checkmark} & \multirow{2}{*}[0.1em]{-} & Uncertainty Masks in Prototype Learning & \multirow{2}{*}[0.1em]{Segmentation} & \multirow{2}{*}[0.1em]{TBAD}\\[3pt]
    
    \midrule 
    \multirow{5}{*}[0.4em]{\citep{jahmunah_uncertainty_2023}} & \multirow{5}{*}[0.4em]{-} & \multirow{5}{*}[0.4em]{\checkmark} & \multirow{5}{*}[0.4em]{-} & \multirow{5}{*}[0.4em]{\shortstack{Dirichlet Distribution \\ Classifier}}  & \multirow{5}{*}[0.4em]{\shortstack{Time-Series \\ Classification}} & Physikalisch-Technische Bundesanstalt database\\[3pt]

    \midrule \citep{adams2023fully} & \multirow{2}{*}[0.1em]{-} & \multirow{2}{*}[0.1em]{\checkmark} & \multirow{2}{*}[0.1em]{-} & Bayesian Learning \& Variational inference & \multirow{2}{*}[0.1em]{Shape Prediction} & Private Left Atrium Dataset from UUtah$^*$\\[3pt]

    \midrule
    \citep{kazemi2022deue} & \multirow{2}{*}[0.1em]{-}  & \multirow{2}{*}[0.1em]{\checkmark} & \multirow{2}{*}[0.1em]{-} & \multirow{2}{*}[0.1em]{Ensemble Learning} & \multirow{2}{*}[0.1em]{Regression} & \multirow{2}{*}[0.1em]{EchoNet-Dynamic}\\[3pt]

    \midrule
    \multirow{2}{*}[0.1em]{\citep{zhang_heteroscedastic_2024}} & \multirow{2}{*}[0.1em]{-} & \multirow{2}{*}[0.1em]{\checkmark} & \multirow{2}{*}[0.1em]{-} & Displacement and Variance Estimators & \multirow{2}{*}[0.1em]{Segmentation} & ACDC, CAMUS, Private 3D Echo$^*$\\[3pt]

    \midrule
    \citep{barandas_evaluation_2024} & \multirow{2}{*}[0.1em]{-} & \multirow{2}{*}[0.1em]{\checkmark} & \multirow{2}{*}[0.1em]{\checkmark} & Monte Carlo Dropout \& Ensemble Learning & \multirow{2}{*}[0.1em]{Classification} & PhysioNet/CinC Challenge 2020 \\[3pt]

    \midrule
    \multirow{3}{*}[0.1em]{\citep{vranken_uncertainty_2021}} & \multirow{3}{*}[0.1em]{-} & \multirow{3}{*}[0.1em]{-} & \multirow{3}{*}[0.1em]{\checkmark} & Monte Carlo Dropout \& Variational Inference & \multirow{3}{*}[0.1em]{\shortstack{Time-Series \\ Classification}} & CPSC2018-Dynamic, UMCU-Triage$^*$, UMCU-Diagnose$^*$\\[3pt]
    
    %%%%%%%%%%%%%%% NEUROLOGY %%%%%%%%%%%%%%% 
    \midrule
    \multicolumn{7}{c}{\textbf{Neurology}} \\

    \midrule 
    \multirow{2}{*}[0.1em]{\citep{tanno_uncertainty_2019}} & \multirow{2}{*}[0.1em]{\checkmark} & \multirow{2}{*}[0.1em]{\checkmark} & \multirow{2}{*}[0.1em]{-} & Heteroscedastic Noise \& Variational Dropout & \multirow{2}{*}[0.1em]{Image Enhancement} & WU-Minn HCP, Lifespan, Prisma \textdagger\\[3pt]
    
    %%%%%%%%%%%%%%% GENERAL SURGERY %%%%%%%%%%%%%%% 
    \midrule
    \multicolumn{7}{c}{\textbf{General Surgery}} \\

    \midrule
    \multirow{2}{*}[0.1em]{\citep{yang2023uncertainty}} & \multirow{2}{*}[0.1em]{-}  & \multirow{2}{*}[0.1em]{\checkmark} & \multirow{2}{*}[0.1em]{\checkmark} & \multirow{2}{*}[0.1em]{Evidential Learning} & \multirow{2}{*}[0.1em]{Segmentation} & CholecSeg8K, Private Endoscopy Dataset$^*$\\[3pt]

    %%%%%%%%%%%%%%% OPHTHALMOLOGY %%%%%%%%%%%%%%% 
    \midrule
    \multicolumn{7}{c}{\textbf{Ophthalmology}} \\
    
    \midrule 
    \multirow{2}{*}[0.1em]{\citep{hu2021learning}} & \multirow{2}{*}[0.1em]{\checkmark} & \multirow{2}{*}[0.1em]{-} & \multirow{2}{*}[0.1em]{\checkmark}  & Calibration Error Estimation & \multirow{2}{*}[0.1em]{Classification} & \multirow{2}{*}[0.1em]{EyePACS, FastMRI}\\[3pt]

    \midrule \multirow{2}{*}[0.1em]{\citep{ren2023uncertaintyinformedmutuallearningjoint}} & \multirow{2}{*}[0.1em]{-} & \multirow{2}{*}[0.1em]{\checkmark} & \multirow{2}{*}[0.1em]{-} & Evidential Deep Learning & Classification \& Segmentation & \multirow{2}{*}[0.1em]{REFUGE, ISPY I}\\[3pt]
    
    \midrule 
    \multirow{3}{*}[0.1em]{\citep{leibig_leveraging_2016}} & \multirow{3}{*}[0.1em]{-} & \multirow{3}{*}[0.1em]{\checkmark} & \multirow{3}{*}[0.1em]{-} & \multirow{3}{*}[0.1em]{Monte Carlo Dropout} & \multirow{3}{*}[0.1em]{Classification} & Kaggle Diabetic Retinopathy Detection Dataset\\[3pt]
    
    \midrule
    \citep{wu2022nonlinear} & \multirow{2}{*}[0.1em]{-}  & \multirow{2}{*}[0.1em]{\checkmark} & \multirow{2}{*}[0.1em]{\checkmark}  & Ensemble \& Bayesian Learning & Surgery Time Estimation & \multirow{2}{*}[0.1em]{Cataract-101}\\[3pt]

    \midrule    
    \citep{band2022benchmarking} & -  & \checkmark & \checkmark  & Bayesian Learning & Classification & EyePACS, APTOS\\[3pt]

    %%%%%%%%%%%%%%% ORTHOPEDICS %%%%%%%%%%%%%%% 
    \midrule
    \multicolumn{7}{c}{\textbf{Orthopedics}} \\

    \midrule
    \citep{browning_uncertainty_2021} & -  & \checkmark & -  & Q-Learning & Detection & Private Knee MRI$^*$\\[3pt]

    \midrule \citep{wahlstrand_skarstrom_explainable_2024} & \multirow{2}{*}[0.1em]{-} & \multirow{2}{*}[0.1em]{\checkmark} & \multirow{2}{*}[0.1em]{\checkmark} & \multirow{2}{*}[0.1em]{Likelihood Scores} & Fracture Analysis \& Classification & \multirow{2}{*}[0.1em]{SUPERB}\\[3pt]
    
    \midrule \citep{teichmann2024towards} & \multirow{2}{*}[0.1em]{-} & \multirow{2}{*}[0.1em]{\checkmark} & \multirow{2}{*}[0.1em]{\checkmark} & \multirow{2}{*}[0.1em]{Monte Carlo Dropout}  & \multirow{2}{*}[0.1em]{Segmentation} & Private Organs Dataset$^*$\\[3pt]

    % %%%%%%%%%%%%%%% ONCOLOGY %%%%%%%%%%%%%%% 
    \midrule
    \multicolumn{7}{c}{\textbf{Oncology}} \\

    \midrule 
    \multirow{2}{*}[0.1em]{\citep{li2022ultra}} & \multirow{2}{*}[0.1em]{\checkmark}  & \multirow{2}{*}[0.1em]{\checkmark} & \multirow{2}{*}[0.1em]{-}  & Label Probability Distribution & Tumor Cellularity Scoring & \multirow{2}{*}[0.1em]{BreastPathQ}\\[3pt]
    
    \midrule \citep{aljuhani_uncertainty_2022} & - & \checkmark & - & Monte Carlo Dropout & Classification & TCGA\\[3pt]

    \midrule
    \multirow{3}{*}[0.1em]{\citep{luo2021efficient}} & \multirow{3}{*}[0.1em]{-}  & \multirow{3}{*}[0.1em]{\checkmark} & \multirow{3}{*}[0.1em]{-}  & \multirow{3}{*}[0.1em]{\shortstack{Rectified Pyramid \\ Consistency}}  & \multirow{3}{*}[0.1em]{Segmentation} & Private Nasopharyngeal Carcinoma$^*$\\[3pt]

    \midrule
    \citep{zou2022tbrats} & -  & \checkmark & -  & Logic Theory & Segmentation & BraTS\\[3pt]

    \midrule 
    \multirow{2}{*}[0.1em]{\citep{hung_cross-slice_2024}} & \multirow{2}{*}[0.1em]{-} & \multirow{2}{*}[0.1em]{\checkmark} & \multirow{2}{*}[0.1em]{-} & Evidential Deep Learning & \multirow{2}{*}[0.1em]{Classification} & \multirow{2}{*}[0.1em]{PICAI} \\[3pt]

    \midrule 
    \multirow{3}{*}[0.1em]{\citep{zepf_laplacian_2024}} & \multirow{3}{*}[0.1em]{-} & \multirow{3}{*}[0.1em]{\checkmark} & \multirow{3}{*}[0.1em]{-} & \multirow{3}{*}[0.1em]{\shortstack{Laplacian\\ Segmentation Network}}  & \multirow{3}{*}[0.1em]{Segmentation} & ClinSkin, PAD-UFES-20, QUBIQ 2021 \textdagger \\[3pt]

    \midrule 
    \multirow{2}{*}[0.1em]{\citep{dong2024uncertainty}} & \multirow{2}{*}[0.1em]{-} & \multirow{2}{*}[0.1em]{\checkmark} & \multirow{2}{*}[0.1em]{-} & Evidential Deep Learning & \multirow{2}{*}[0.1em]{Image Grading} & Private Prostatic Cancer Dataset$^*$\\[3pt]

    \midrule \citep{ren_skincon_2024} & - & \checkmark & - & Conformal Prediction & Classification & SkinCON \\[3pt]

    \midrule \citep{thiagarajan_explanation_2022} & \multirow{2}{*}[0.1em]{-} & \multirow{2}{*}[0.1em]{\checkmark} & \multirow{2}{*}[0.1em]{-} & \multirow{2}{*}[0.1em]{Bayesian Learning} & \multirow{2}{*}[0.1em]{Classification} & Breast Histopathology Kaggle\\[3pt]

    \midrule \citep{schott_uncertainty_2024} & - & \checkmark & - & Localized Gradients & Segmentation & LiTS, Bone Metastates\\[3pt]

    \midrule 
    \multirow{3}{*}[0.1em]{\citep{buddenkotte_calibrating_2023}} & \multirow{3}{*}[0.1em]{-} & \multirow{3}{*}[0.1em]{\checkmark} & \multirow{3}{*}[0.1em]{-} & \multirow{3}{*}[0.1em]{\shortstack{Bayesian \& Ensemble \\ Learning}} & \multirow{3}{*}[0.1em]{Segmentation} & KiTS19, Private Ovarian Cancer CT Scans$^*$\\[3pt]

    \midrule \citep{hu_uncertainty_2021} & - & \checkmark & - & Gaussian Process & Radiogenomics EGFR amplification & Private Self-Recorded Data$^*$\\[3pt]

    \midrule 
    \multirow{2}{*}[0.1em]{\citep{abdar_uncertainty_2021}} & \multirow{2}{*}[0.1em]{-} & \multirow{2}{*}[0.1em]{\checkmark} & \multirow{2}{*}[0.1em]{-} & Monte Carlo Dropout \& Ensemble Learning & \multirow{2}{*}[0.1em]{Classification} & \multirow{2}{*}[0.1em]{ISIC 2019, HAM10000}\\[3pt]

    \midrule 
    \citep{zhou_uncertainty_2023} & - & \checkmark & - & Monte Carlo Dropout & Segmentation & BraTS 2018 \& 2019\\[3pt]

    \midrule 
    \citep{luo_uncertainty-guided_2024} & \multirow{2}{*}[0.1em]{-} & \multirow{2}{*}[0.1em]{\checkmark} & \multirow{2}{*}[0.1em]{-} & Active Selection Sampling & \multirow{2}{*}[0.1em]{Segmentation} & \multirow{2}{*}[0.1em]{DPL, FSM, UPL}\\[3pt]

    \midrule 
    \multirow{3}{*}[0.1em]{\citep{sahlsten_application_2024}} & \multirow{3}{*}[0.1em]{-} & \multirow{3}{*}[0.1em]{\checkmark} & \multirow{3}{*}[0.1em]{\checkmark} & \multirow{3}{*}[0.1em]{Bayesian Learning} & \multirow{3}{*}[0.1em]{Segmentation} & HECKTOR, Private U-Texas Cancer Center$^*$\\[3pt]
    
    \midrule 
    \multirow{2}{*}[0.1em]{\citep{peluso_deep_2024}} & \multirow{2}{*}[0.1em]{-} & \multirow{2}{*}[0.1em]{-} & \multirow{2}{*}[0.1em]{\checkmark} & Deep Abstaining Classifier & Clinical Text Classification & \multirow{2}{*}[0.1em]{NCI SEER Report}\\[3pt]

    \midrule 
    \citep{hamedani-karazmoudehfar_breast_2023} & \multirow{3}{*}[0.1em]{-} & \multirow{3}{*}[0.1em]{-} & \multirow{3}{*}[0.1em]{\checkmark} & \multirow{3}{*}[0.1em]{\shortstack{Monte Carlo Dropout \\ \& Ensemble Learning}} & \multirow{3}{*}[0.1em]{Classification} & \multirow{3}{*}[0.1em]{WDBC}\\[3pt]

    \midrule 
    \citep{ehr_1} & \multirow{3}{*}[0.1em]{-} & \multirow{3}{*}[0.1em]{\checkmark} & \multirow{3}{*}[0.1em]{\checkmark} & Bayesian Gaussian-Orocess-based UQ Framework & \multirow{3}{*}[0.1em]{CATE Estimation} & \multirow{3}{*}[0.1em]{Synthetic, ACIC}\\[3pt]

    \midrule 
    \multirow{2}{*}[0.1em]{\citep{ehr_2}} & \multirow{2}{*}[0.1em]{-} & \multirow{2}{*}[0.1em]{\checkmark} & \multirow{2}{*}[0.1em]{\checkmark} & Approximate Bayesian UQ & \multirow{2}{*}[0.1em]{CATE Estimation} & \multirow{2}{*}[0.1em]{Simulated, MIMIC-IV}\\[3pt]

    \midrule 
    \multirow{2}{*}[0.1em]{\citep{ehr_3}} & \multirow{2}{*}[0.1em]{-} & \multirow{2}{*}[0.1em]{-} & \multirow{2}{*}[0.1em]{\checkmark} & \multirow{2}{*}[0.1em]{Monte Carlo} & \multirow{2}{*}[0.1em]{CATE Estimation} & CVSim, Cancer Growth\\[3pt]

    \midrule 
    \multirow{5}{*}[0.4em]{\citep{ehr_4}} & \multirow{5}{*}[0.4em]{-} & \multirow{5}{*}[0.4em]{\checkmark} & \multirow{5}{*}[0.4em]{\checkmark} & \multirow{5}{*}[0.4em]{\shortstack{Bayesian Neural \\ Controlled Differential \\ Equation}} & \multirow{5}{*}[0.4em]{CATE Estimation} & Simulated, Pharmacokinetic-pharmacodynamic Tumor Growth Model\\[3pt]

    \midrule 
    \multirow{6}{*}[0.4em]{\citep{ehr_5}} & \multirow{6}{*}[0.4em]{-} & \multirow{6}{*}[0.4em]{\checkmark} & \multirow{6}{*}[0.4em]{\checkmark} & \multirow{6}{*}[0.4em]{\shortstack{Uncertainty-Aware\\ Latent Neural ODE}} & \multirow{6}{*}[0.4em]{\shortstack{Individualized\\treatment effect \\ Estimation}} & Synthetic, Cardiovascular System Modeling, Pharmacodynamics Model\\[3pt]

    \midrule 
    \multirow{5}{*}[0.4em]{\citep{ehr_6}} & \multirow{5}{*}[0.4em]{-} & \multirow{5}{*}[0.4em]{\checkmark} & \multirow{5}{*}[0.4em]{\checkmark} & \multirow{5}{*}[0.4em]{\shortstack{Evidential Regression\\ Network}}   & \multirow{5}{*}[0.4em]{\shortstack{Time-to-Event\\ Prediction}}  & Synthetic, Simulated, METABRIC, GBSG, SUPPORT, MIMIC-IV\\[3pt]

    % %%%%%%%%%%%%%%% Pulmonology %%%%%%%%%%%%%%% 
    \midrule
    \multicolumn{7}{c}{\textbf{Pulmonology}} \\

    \midrule 
    \multirow{2}{*}[0.1em]{\citep{li_dynamic_2023}} & \multirow{2}{*}[0.1em]{-} & \multirow{2}{*}[0.1em]{\checkmark} & \multirow{2}{*}[0.1em]{-} & Dirichlet Distribution Classifier & \multirow{2}{*}[0.1em]{Classification} & \multirow{2}{*}[0.1em]{ISIC18, Chest XRay8} \\[3pt]

    \midrule 
    \multirow{2}{*}[0.1em]{\citep{yang2022uncertainty}} & \multirow{2}{*}[0.1em]{-} & \multirow{2}{*}[0.1em]{\checkmark} & \multirow{2}{*}[0.1em]{-}  & Attention Masks for Uncertainty & \multirow{2}{*}[0.1em]{Segmentation} & \multirow{2}{*}[0.1em]{LIDC-IDRI} \\[3pt]

      % %%%%%%%%%%%%%%% SPECIALIZED APPLICATIONS %%%%%%%%%%%%%%% 
    \midrule
    \multicolumn{7}{c}{\textbf{Specialized Applications}} \\

    \midrule 
    \multirow{2}{*}[0.1em]{\citep{ji_unraveling_2024}} & \multirow{2}{*}[0.1em]{-} & \multirow{2}{*}[0.1em]{\checkmark} & \multirow{2}{*}[0.1em]{-} & PCA Based Uncertainty Weighting & \multirow{2}{*}[0.1em]{Pain Assessment} & Biovid, Private Apon Dataset$^*$\\[3pt]

    \midrule
    \multirow{2}{*}[0.1em]{\citep{kang_statistical_2021}} & \multirow{2}{*}[0.1em]{-} & \multirow{2}{*}[0.1em]{\checkmark} & \multirow{2}{*}[0.1em]{-} & \multirow{2}{*}[0.1em]{Shannon Entropy} & Sleep Pattern Assessment & Private Sleep Pattern Dataset$^*$\\[3pt]

    \midrule
    \multirow{3}{*}[0.1em]{\citep{lu_uncertainty_2024}} & \multirow{3}{*}[0.1em]{-} & \multirow{3}{*}[0.1em]{\checkmark} & \multirow{3}{*}[0.1em]{-} & \multirow{3}{*}[0.1em]{\shortstack{Hierarchical \\ Interaction Network}}  & \multirow{3}{*}[0.1em]{\shortstack{Clinical Trial \\ Approval Prediction}}
    & TOP clinical Trial Approval Prediction Benchmark\\[3pt]

    \midrule 
    \multirow{2}{*}[0.1em]{\citep{dusenberry_analyzing_2020}} & \multirow{2}{*}[0.1em]{-} & \multirow{2}{*}[0.1em]{\checkmark} & \multirow{2}{*}[0.1em]{-} & Bayesian \& Ensemble Learning & \multirow{2}{*}[0.1em]{Intensive Care Unit} & \multirow{2}{*}[0.1em]{MIMIC-III, eICU}\\[3pt]

    \midrule 
    \multirow{3}{*}[0.1em]{\citep{oszkinat_uncertainty_2023}} & \multirow{3}{*}[0.1em]{-} & \multirow{3}{*}[0.1em]{\checkmark} & \multirow{3}{*}[0.1em]{-} & \multirow{3}{*}[0.1em]{\shortstack{Residual-Augmented \\ Loss Function}}  & Blood Alcohol Concentration Estimation & \multirow{3}{*}[0.1em]{\shortstack{Alcohol Concentration \\ Data}}\\[3pt]

    \midrule 
    \multirow{3}{*}[0.1em]{\citep{jeong2024uncertainty}} & \multirow{3}{*}[0.1em]{-} & \multirow{3}{*}[0.1em]{\checkmark} & \multirow{3}{*}[0.1em]{-} & Pixel-Wise Uncertainty for Diffusion Model & \multirow{3}{*}[0.1em]{\shortstack{Image Generation  \& \\ Adversarial Attacks}} & \multirow{3}{*}[0.1em]{\shortstack{Kvasir-SEG,\\ ETIS-Larib Polyp DB}} \\[3pt]
    
    \midrule 
    \citep{li_deep_2021} & - & \checkmark & \checkmark & Gaussian Processes & Classification & CPRD\\[3pt]
    
    \midrule 
    \multirow{2}{*}[0.1em]{\citep{konuk2024framework}} & \multirow{2}{*}[0.1em]{-} & \multirow{2}{*}[0.1em]{\checkmark} & \multirow{2}{*}[0.1em]{\checkmark} & Entropy \& Confidence Based Uncertainty & \multirow{2}{*}[0.1em]{Classification} & \multirow{2}{*}[0.1em]{Private OMLC-RS$^*$}\\[3pt]

    \midrule 
    \citep{durso-finley_improving_2023} &  \multirow{3}{*}[0.1em]{-} & \multirow{3}{*}[0.1em]{-} & \multirow{3}{*}[0.1em]{\checkmark} & \multirow{3}{*}[0.1em]{\shortstack{Bayesian Causal \\ Models}} & Factual Error Correlation with Uncertainty & \multirow{3}{*}[0.1em]{\shortstack{BRAVO, OPERA 1-2, \\ DEFINE \textdagger}}\\[3pt]
    \midrule
    \newpage\\[-5pt]
    %%%%%%%%%%%%%%% MEDICAL IMAGING %%%%%%%%%%%%%%%
    % different to RADIOLOGY, but groups studies that evaluate on different imaging datasets
    \multicolumn{7}{c}{\textbf{Medical Imaging}} \\

    \midrule
    \multirow{2}{*}[0.1em]{\citep{lee2023diffusion}} & \multirow{2}{*}[0.1em]{\checkmark} & \multirow{2}{*}[0.1em]{-} & \multirow{2}{*}[0.1em]{-}  & Diffusion Probabilistic Modeling & \multirow{2}{*}[0.1em]{Noise Reduction} & \multirow{2}{*}[0.1em]{Private MR Dataset$^*$}\\[3pt]
    
    \midrule
    \multirow{2}{*}[0.1em]{\citep{khader2023denoising}} & \multirow{2}{*}[0.1em]{\checkmark} & \multirow{2}{*}[0.1em]{-} & \multirow{2}{*}[0.1em]{-} & Diffusion Probabilistic Modeling & \multirow{2}{*}[0.1em]{Data Generation} & \multirow{2}{*}[0.1em]{ADNI, Breast MRI}\\[3pt]
    
    \midrule
    \multirow{2}{*}[0.1em]{\citep{iuliano2024denoising}} & \multirow{2}{*}[0.1em]{\checkmark} & \multirow{2}{*}[0.1em]{\checkmark} & \multirow{2}{*}[0.1em]{-} & Diffusion Probabilistic Modeling & \multirow{2}{*}[0.1em]{Data Generation} & \multirow{2}{*}[0.1em]{NLM Malaria}\\[3pt]

    \midrule
    \multirow{2}{*}[0.1em]{\citep{adib2023synthetic}} & \multirow{2}{*}[0.1em]{\checkmark} & \multirow{2}{*}[0.1em]{\checkmark} & \multirow{2}{*}[0.1em]{-}  & Diffusion Probabilistic Modeling & \multirow{2}{*}[0.1em]{Data Generation} & \multirow{2}{*}[0.1em]{MIT-BIH Arrhythmia}\\[3pt]
    
    \midrule
    \citep{angelopoulos2022image} & \multirow{2}{*}[0.1em]{\checkmark}  & \multirow{2}{*}[0.1em]{\checkmark} & \multirow{2}{*}[0.1em]{-}  & Pixel-wise Uncertainty Intervals & \multirow{2}{*}[0.1em]{Segmentation} & BSCCM, TEMCA2, FastMRI\\[3pt]
    
    \midrule 
    \multirow{3}{*}[0.1em]{\citep{das_confidence-guided_2024}} & \multirow{3}{*}[0.1em]{\checkmark} & \multirow{3}{*}[0.1em]{\checkmark} & \multirow{3}{*}[0.1em]{-} & Confidence Guided Pseudo-Label Optimizer & \multirow{3}{*}[0.1em]{Segmentation} & \multirow{3}{*}[0.1em]{\shortstack{VinDr-CXR,\\ TBX11K, B-Fract}}\\[3pt]

    \midrule 
    \multirow{3}{*}[0.1em]{\citep{scalco_uncertainty_2024}} & \multirow{3}{*}[0.1em]{-} & \multirow{3}{*}[0.1em]{\checkmark} & \multirow{3}{*}[0.1em]{-} & \multirow{3}{*}[0.1em]{\shortstack{Bayesian \& Ensemble\\ Learning}} & \multirow{3}{*}[0.1em]{Segmentation} & Kidney Tumor Segmentation Challenge 2021\\[3pt]

    \midrule 
    \multirow{2}{*}[0.1em]{\citep{judge_asymmetric_2023}} & \multirow{2}{*}[0.1em]{-} & \multirow{2}{*}[0.1em]{\checkmark} & \multirow{2}{*}[0.1em]{-} & Gaussian Probability Distributions & \multirow{2}{*}[0.1em]{Segmentation} & CAMUS, JSRT, Private US Dataset$^*$ \\[3pt]
    
    \midrule \citep{lennartz_segmentation_2023} & \multirow{2}{*}[0.1em]{-} & \multirow{2}{*}[0.1em]{\checkmark} & \multirow{2}{*}[0.1em]{-} & Distance Regularization & \multirow{2}{*}[0.1em]{Segmentation} & Calgary-Campinas-359, ACDC, M\&MS \\[3pt]

    \midrule \citep{larrazabal_maximum_2023} & \multirow{2}{*}[0.1em]{-} & \multirow{2}{*}[0.1em]{\checkmark} & \multirow{2}{*}[0.1em]{-} & \multirow{2}{*}[0.1em]{KL Divergence} & \multirow{2}{*}[0.1em]{Segmentation} & Atrial Segmentation Challenge, WMH\\[3pt]
    
    \midrule
    \multirow{3}{*}[0.1em]{\citep{xiang2022fussnet}} & \multirow{3}{*}[0.1em]{-}  & \multirow{3}{*}[0.1em]{\checkmark} & \multirow{3}{*}[0.1em]{-}  & \multirow{3}{*}[0.1em]{Unsupervised Learning} & \multirow{3}{*}[0.1em]{Segmentation} & DeepOrgan, 2018 Atria Segmentation Challenge\\[3pt]

    \midrule
    \multirow{2}{*}[0.1em]{\citep{judge_crisp_2022}} & \multirow{2}{*}[0.1em]{-}  & \multirow{2}{*}[0.1em]{\checkmark} & \multirow{2}{*}[0.1em]{-} & \multirow{2}{*}[0.1em]{Contrastive Learning} & \multirow{2}{*}[0.1em]{Segmentation} & CAMUS, HMC-QU, Shenzen \textdagger\\[3pt]

    \midrule
    \multirow{2}{*}[0.1em]{\citep{xie2022uncertainty}} & \multirow{2}{*}[0.1em]{-} & \multirow{2}{*}[0.1em]{\checkmark} & \multirow{2}{*}[0.1em]{-} & Uncertainty Attention Module & \multirow{2}{*}[0.1em]{Segmentation} & CAMUS, TN-SCUI, HC18 \\[3pt]

    \midrule 
    \multirow{2}{*}[0.1em]{\citep{fu2025cam}} & \multirow{2}{*}[0.1em]{-} & \multirow{2}{*}[0.1em]{\checkmark} & \multirow{2}{*}[0.1em]{-} & Uncertainty Weighted Class Activation Maps & \multirow{2}{*}[0.1em]{Segmentation} & Private Fetal Brain Dataset$^*$\\[3pt]

    \midrule 
    \multirow{3}{*}[0.1em]{\citep{abdar_hercules_2023}} & \multirow{3}{*}[0.1em]{-} & \multirow{3}{*}[0.1em]{\checkmark} & \multirow{3}{*}[0.1em]{-} & \multirow{3}{*}[0.1em]{Monte Carlo Dropout} & \multirow{3}{*}[0.1em]{Classification} & Retinal OCT, Lung CT, Pneumonia Chest X-Ray\\[3pt]

    \midrule
    \citep{upadhyay2021uncertainty} & \multirow{2}{*}[0.1em]{-}  & \multirow{2}{*}[0.1em]{\checkmark} & \multirow{2}{*}[0.1em]{-}  & Uncertainty-Guided GAN & \multirow{2}{*}[0.1em]{Classification} & IXI, Private PET to CT Dataset$^*$\\[3pt]
    
    \midrule 
    \multirow{2}{*}[0.1em]{\citep{sharma_confidence_2024}} & \multirow{2}{*}[0.1em]{-} & \multirow{2}{*}[0.1em]{\checkmark} & \multirow{2}{*}[0.1em]{-} & Entropy Driven Distillation Learning & \multirow{2}{*}[0.1em]{Classification} & \multirow{2}{*}[0.1em]{HAM10000, APTOS} \\[3pt]
    
    \midrule 
    \multirow{2}{*}[0.1em]{\citep{qendro_early_nodate}} & \multirow{2}{*}[0.1em]{-} & \multirow{2}{*}[0.1em]{\checkmark} & \multirow{2}{*}[0.1em]{-} & \multirow{2}{*}[0.1em]{Ensemble Learning} & Classification Benchmarking & ECG5000, EEG, ISIC2018\\[3pt]

    \midrule 
    \multirow{2}{*}[0.1em]{\citep{abboud_sparse_2024}} & \multirow{2}{*}[0.1em]{-} & \multirow{2}{*}[0.1em]{\checkmark} & \multirow{2}{*}[0.1em]{-} & \multirow{2}{*}[0.1em]{Bayesian Learning} & Classification \& Segmentation & ISIC, ChestMNIST, LIDC-IDRI\\[3pt]

    \midrule \citep{samareh_uq-chi_2019} & - & \checkmark & - & Contemporaneous Longitudinal Index & Degenerative Disease Modeling & Private Alzheimer's Dataset$^*$\\[3pt]

    \midrule 
    \multirow{3}{*}[0.1em]{\citep{ramesh_geometric_2024}} & \multirow{3}{*}[0.1em]{-} & \multirow{3}{*}[0.1em]{\checkmark} & \multirow{3}{*}[0.1em]{-} & \multirow{3}{*}[0.1em]{\shortstack{Multi-head Geometric \\ Transformations}} & \multirow{3}{*}[0.1em]{3D Pose Prediction} & INTERGROWTH Fetal Brain Ultrasound\\[3pt]
    
    \midrule 
    \multirow{3}{*}[0.1em]{\citep{chen2024modeling}} & \multirow{3}{*}[0.1em]{-} & \multirow{3}{*}[0.1em]{\checkmark} & \multirow{3}{*}[0.1em]{\checkmark} & \multirow{3}{*}[0.1em]{Conformal Prediction} & \multirow{3}{*}[0.1em]{\shortstack{Classification \&\\ Segmentation}} & ISIC 2018, BloodMNIST, OrganCMNIST\\[3pt]
    
    \midrule 
    \multirow{2}{*}[0.1em]{\citep{gong_patient-specific_2023}} & \multirow{2}{*}[0.1em]{-} & \multirow{2}{*}[0.1em]{\checkmark} & \multirow{2}{*}[0.1em]{\checkmark} & Bayesian Learning \& Knowledge Distillation & \multirow{2}{*}[0.1em]{Image Denoising} & \multirow{2}{*}[0.1em]{Low-Dose CT Image}\\[3pt]

    \midrule
    \citep{popordanoska2021relationship} & \multirow{2}{*}[0.1em]{-} & \multirow{2}{*}[0.1em]{\checkmark} & \multirow{2}{*}[0.1em]{\checkmark} & \multirow{2}{*}[0.1em]{Model Calibration} & \multirow{2}{*}[0.1em]{Segmentation} & \multirow{2}{*}[0.1em]{BraTS, ISLES}\\[3pt]
    
    \midrule
    \citep{kushibar2022layer} & -  & \checkmark & \checkmark  & Ensemble Learning & Segmentation & BCDR, MnM\\[3pt]

    \midrule 
    \multirow{3}{*}[0.1em]{\citep{shi_evidential_2024}} & \multirow{3}{*}[0.1em]{-} & \multirow{3}{*}[0.1em]{\checkmark} & \multirow{3}{*}[0.1em]{\checkmark} & \multirow{3}{*}[0.1em]{\shortstack{Dempster-Shafer \\ Theory}} & \multirow{3}{*}[0.1em]{Segmentation} & MitoEM-(R,H), Kasthuri++, Lucchi++ \textdagger \\[3pt]

    \midrule 
    \multirow{3}{*}[0.1em]{\citep{gu_revisiting_2024}} & \multirow{3}{*}[0.1em]{-} & \multirow{3}{*}[0.1em]{\checkmark} & \multirow{3}{*}[0.1em]{\checkmark} & \multirow{3}{*}[0.1em]{Ensemble Learning} & \multirow{3}{*}[0.1em]{Segmentation} & RSNA Pneumonia, VinDr-CXR, Brain MRI \textdagger\\[3pt]
     
    \midrule
    \multirow{2}{*}[0.1em]{\citep{yang2022uncertainty}} & \multirow{2}{*}[0.1em]{-} & \multirow{2}{*}[0.1em]{\checkmark} & \multirow{2}{*}[0.1em]{\checkmark}  & Uncertainty Attention Masks & \multirow{2}{*}[0.1em]{Segmentation} & \multirow{2}{*}[0.1em]{LIDC-IDRI}\\[3pt]
    
    \midrule
    \multirow{2}{*}[0.1em]{\citep{zhang_uncertainty_2023}} & \multirow{2}{*}[0.1em]{-} & \multirow{2}{*}[0.1em]{\checkmark} & \multirow{2}{*}[0.1em]{\checkmark} & Monte Carlo Dropout \& Ensemble Learning & \multirow{2}{*}[0.1em]{Segmentation} & Heart-Disease, ISIC2019, PAMAP2 \\[3pt]

    \bottomrule\\[3pt]
\multicolumn{7}{l}{* Private datasets for specific medical applications, \textdagger Evaluation on more than 3 datasets.} \\
\multicolumn{7}{l}{Classification and segmentation refer to imaging based tasks unless specified otherwise.} \\

\end{longtable}
}

\end{document}